\documentclass[accepted]{uai2023}
\usepackage[american]{babel}
\usepackage{natbib}
\usepackage{mathtools}
\usepackage{booktabs}
\usepackage{tikz}
\usepackage{hyperref}
\hypersetup{
  colorlinks = true,
  linkcolor=linkblue,   
  citecolor=linkblue,   
  urlcolor=linkblue,    
  bookmarksopen=true,
  pdfdisplaydoctitle=true
}
\usepackage{amsmath}
\usepackage{amsthm}
\usepackage{amssymb}
\theoremstyle{definition}
\newtheorem{definition}{Definition}
\newtheorem{theorem}{Theorem}
\newtheorem{lemma}[theorem]{Lemma}
\newtheorem{assumption}{Assumption}
\usepackage{xcolor}
\usepackage{float}
\usepackage{adjustbox}
\usepackage{bbm}
\usepackage{cancel}
\usepackage{soul}
\usepackage[titlenumbered,ruled]{algorithm2e}

\definecolor{linkblue}{rgb}{0.0, 0.3, 0.6}
\definecolor{dartmouthgreen}{rgb}{0.05, 0.5, 0.06}
\definecolor{frenchblue}{rgb}{0.0, 0.45, 0.73}
\definecolor{mediumred-violet}{rgb}{0.73, 0.2, 0.52}
\definecolor{darkorange}{rgb}{0.80, 0.439, 0}
\definecolor{orange(ryb)}{rgb}{0.98, 0.6, 0.01}
\definecolor{darkorchid}{rgb}{0.6, 0.2, 0.8}
\definecolor{independence}{RGB}{187,212,113}
\definecolor{independence-text}{RGB}{145,176,53} 
\definecolor{weakdependence}{RGB}{99,164,108}
\definecolor{weakdependence-text}{RGB}{68,117,75} 
\definecolor{moderatedependence}{RGB}{35,51,41}
\definecolor{moderatedependence-text}{RGB}{52,76,61} 

\def\prob#1{\Pr(\, #1 \,)}
\def\Cprob#1#2{\prob{ #1 \,|\,#2}}
\def\eg{{\em e.g.},\ }
\def\ie{{\em i.e.},\ }
\long\def\comment#1{}

\newtheorem{manuallemmainner}{Lemma}
\newenvironment{manuallemma}[1]{
  \renewcommand\themanuallemmainner{#1}
  \manuallemmainner
}{\endmanuallemmainner}

\allowdisplaybreaks

\begin{document}

\title{Copula-Based Deep Survival Models for Dependent Censoring}
\author[1,2]{Ali Hossein Gharari Foomani$^{*,}$}
\author[3,5]{Michael Cooper$^{*,\dagger,}$}
\author[1,2]{Russell Greiner}
\author[3,4,5]{Rahul G. Krishnan}
\affil[1]{
Department of Computing Science, University of Alberta
}
\affil[2]{
Alberta Machine Intelligence Institute
}
\affil[3]{
Department of Computer Science, University of Toronto
}
\affil[4]{
Department of Laboratory Medicine and Pathobiology, University of Toronto
}
\affil[5]{
Vector Institute
}
  
\maketitle
\begin{NoHyper}
\def\thefootnote{*}\footnotetext{These authors contributed equally to this work.}\def\thefootnote{\arabic{footnote}}
\end{NoHyper}
\def\thefootnote{$\dagger$}\footnotetext{Correspondence to \href{mailto:coopermj@cs.toronto.edu}{coopermj@cs.toronto.edu}.}\def\thefootnote{\arabic{footnote}}

\begin{abstract}
A survival dataset describes a set of instances (\eg patients) and provides, for each, either the time until an event (\eg death), or the censoring time (\eg when lost to follow-up -- which is a lower bound on the time until the event). We consider the challenge of survival prediction: learning, from such data, a predictive model that can produce an individual survival distribution for a novel instance. Many contemporary methods of survival prediction implicitly assume that the event and censoring distributions are independent conditional on the instance’s covariates –- a strong assumption that is difficult to verify (as we observe only one outcome for each instance) and which can induce significant bias when it does not hold. This paper presents a parametric model of survival that extends modern non-linear survival analysis by relaxing the assumption of conditional independence. On synthetic and semi-synthetic data, our approach significantly improves estimates of survival distributions compared to the standard that assumes conditional independence in the data.\footnotemark
\end{abstract} 
\footnotetext{Code available at \href{https://github.com/rgklab/copula_based_deep_survival}{this GitHub repository}.}

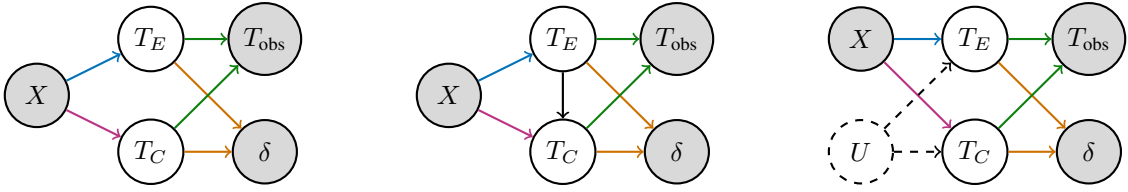
\begin{figure*}
\centering
\begin{tikzpicture}[->, thick]
\node[circle, draw, fill=gray!30, minimum size=24pt] at (0, -0.75) (x) {$X$};
\node[circle, draw, minimum size=24pt] (te) at (1.5, 0) {$T_E$};
\node[circle, draw, minimum size=24pt] (tc) at (1.5, -1.5) {$T_C$};
\node[circle, draw, fill=gray!30, minimum size=24pt] (t) at (3, 0) {$T_{\text{obs}}$};
\node[circle, draw, fill=gray!30, minimum size=24pt] (d) at (3, -1.5) {$\delta$};
  \path[every node/.style={sloped,anchor=south,auto=false}]
        (x) edge             [frenchblue] node {} (te)
        (x) edge             [mediumred-violet] node {} (tc)
        (te) edge            [darkorange] node {} (d)
        (te) edge            [dartmouthgreen] node {} (t)
        (tc) edge            [darkorange] node {} (d)
        (tc) edge            [dartmouthgreen] node {} (t);
\end{tikzpicture}
\qquad\qquad
\begin{tikzpicture}[->, thick]
\node[circle, draw, fill=gray!30, minimum size=24pt] at (0, -0.75) (x) {$X$};
\node[circle, draw, minimum size=24pt] (te) at (1.5, 0) {$T_E$};
\node[circle, draw, minimum size=24pt] (tc) at (1.5, -1.5) {$T_C$};
\node[circle, draw, fill=gray!30, minimum size=24pt] (t) at (3, 0) {$T_{\text{obs}}$};
\node[circle, draw, fill=gray!30, minimum size=24pt] (d) at (3, -1.5) {$\delta$};
  \path[every node/.style={sloped,anchor=south,auto=false}]
        (x) edge             [frenchblue] node {} (te)
        (x) edge             [mediumred-violet] node {} (tc)
        (te) edge            [black] node {} (tc)
        (te) edge            [darkorange] node {} (d)
        (te) edge            [dartmouthgreen] node {} (t)
        (tc) edge            [darkorange] node {} (d)
        (tc) edge            [dartmouthgreen] node {} (t);
\end{tikzpicture}
\qquad\qquad
\begin{tikzpicture}[->, thick]
\node[circle, draw, fill=gray!30, minimum size=24pt] at (0, 0) (x) {$X$};
\node[circle, draw, dashed, minimum size=24pt] at (0, -1.5) (u) {$U$};
\node[circle, draw, minimum size=24pt] (te) at (1.5, 0) {$T_E$};
\node[circle, draw, minimum size=24pt] (tc) at (1.5, -1.5) {$T_C$};
\node[circle, draw, fill=gray!30, minimum size=24pt] (t) at (3, 0) {$T_{\text{obs}}$};
\node[circle, draw, fill=gray!30, minimum size=24pt] (d) at (3, -1.5) {$\delta$};
  \path[every node/.style={sloped,anchor=south,auto=false}]
        (x) edge              [frenchblue] node {} (te)
        (x) edge              [mediumred-violet] node {} (tc)
        (u) edge              [black, dashed] node {} (te)
        (u) edge              [black, dashed] node {} (tc)
        (te) edge             [darkorange] node {} (d)
        (te) edge             [dartmouthgreen] node {} (t)
        (tc) edge             [darkorange] node {} (d)
        (tc) edge             [dartmouthgreen] node {} (t);
\end{tikzpicture}
\caption{\small Three graphical models of survival analysis, showcasing the dependencies between covariates $X$, event/censorship times $T_E$ / $T_C$, time of last observation $T_{obs}$ and event indicator $\delta$. Shaded nodes represent variables whose values we can observe. Blue and magenta arrows represent the event and censoring functions \textcolor{frenchblue}{$f_E : \mathcal{X} \rightarrow \mathbb{R}_+$}, \textcolor{mediumred-violet}{$f_C : \mathcal{X} \rightarrow \mathbb{R}_+$}, respectively, of arbitrary functional form. Green arrows into a node $T$ represent the function \textcolor{dartmouthgreen}{$\mathbb{R}_+^2 \rightarrow \mathbb{R}_+$} defined by\textcolor{dartmouthgreen}{$\min\left(t_e, t_c\right)$}. Orange arrows into a node $\delta$ represent the indicator function \textcolor{darkorange}{$\mathbb{R}_+^2 \rightarrow \{0, 1\}$} defined by \textcolor{darkorange}{$\mathbbm{1}\left[t_e < t_c\right]$}. The leftmost graph demonstrates the case of conditionally independent censoring (or censoring-at-random, CAR), because conditioning on $X$ $d$-separates~\citep{geiger1990d} $T_E$ and $T_C$. The center- and right-most graphs represent cases in which the censoring and event times may be conditionally dependent (or censoring-not-at-random, CNAR): in the center graph, this is through a direct dependency between $T_E$ and $T_C$, while in the rightmost graph, this is via the unobserved confounding node, $U$, that affects both $T_E$ and $T_C$.}
\label{fig:survival-graphs}
\end{figure*}

\addtocontents{toc}{\protect\setcounter{tocdepth}{0}}
\section{Introduction} 
Clinical and epidemiological investigations often want to predict the time until the onset of an event of interest. As examples, a clinical trial of a therapeutic cancer regimen may compare the time-to-mortality in patients who received an experimental therapy against the times of the patients in the control arm~\citep{emmerson2021understanding, zhang2011antiangiogenic}; and a study developing a clinical risk score may want to regress the time until patient mortality onto covariates of interest, in order to leverage the learned model parameters in a predictive risk algorithm~\citep{jia2019cox}.

In such time-to-event prediction tasks, it is common to only have a lower bound on the time-to-event for some instances in the study cohort. Here, we focus on \textit{right censored} instances -- \eg patients who left the study prior to their time of death (loss-to-follow-up), or patients who did not die prior to the conclusion of the study (administrative censoring)~\citep{leung1997censoring, lesko2018censor}. \textit{Survival prediction} refers to the development of statistical models that support time-to-event prediction when some training instances are censored. Rather than discarding such censored instances, methods in survival analysis instead leverage the censoring time as a \textit{lower bound} on that individual's time-to-event~\citep{kalbfleisch2011statistical}.

Let $X^{(i)} \in \mathcal{X}$ refer to the covariates of the $i^{th}$ patient, and let $T_{\text{obs}}^{(i)} \in \mathbb{R}_+$ refer to their time of last observation, taken to be the minimum of the event time $T_E^{(i)} \in \mathbb{R}_+$ and censorship time $T_C^{(i)} \in \mathbb{R}_+$. Because a patient can be either censored or uncensored, but not both, we only observe one of $\{\,T_E, \, T_C\,\}$ for each patient. A common assumption in survival analysis is \textit{conditionally independent censoring}~\citep{kalbfleisch2011statistical}: 
\begin{equation}
T_E\ \perp\ T_C \ |\ X
\label{eq:conditional_indep}
\end{equation}
\ie once $X$ is known, knowing either the event or censoring time does not provide additional information about the other quantity; see Figure~\hbox{\ref{fig:survival-graphs}}(left). This assumption does not always hold. Figure~\ref{fig:survival-graphs} shows this assumption is violated when the event time affects the censoring time, or in the presence of unobserved confounding variables. When Equation~\ref{eq:conditional_indep} does not hold, we say that the data features \textit{dependent censorship}, a common feature of survival data that is unaccounted, or assumed to be absent, in modern survival prediction.

This is not a theoretical concern. Consider a study assessing the survival outcomes of a cohort of chronic disease patients treated with a certain type of medication. The study collects basic demographic and medical information about each patient, their time-of-death or censorship, and an indicator expressing whether the patient died or was censored.

Now imagine that sicker patients often remove themselves from the study in order to explore alternative treatment options. This presents a form of selection bias: while we may surmise that a patient who is censored is more likely to be sicker than their uncensored counterpart, and therefore, may have a lower time-of-death, a statistical model that does not account for this will likely over-estimate each patient's survival time, which may have implications when assessing the safety and utility of the medication in question. This motivating example, characterized by the middle graph in Figure \ref{fig:survival-graphs}, presents a scenario that contemporary approaches to survival regression are often poorly equipped to accommodate. 

Note that it is typically impossible to verify a dependency in practice because we only observe one outcome (either event or censorship) per instance, 
but never both. Also, this dependency between $T_E$ and $T_C$ can be quite subtle if it takes place by means of unobserved confounding variables. The effect of variables $U$ are highlighted in Figure~\mbox{\ref{fig:survival-graphs}} (right).

Relaxing the conditional independence assumption of Equation~\ref{eq:conditional_indep} has been previously studied. However, existing approaches either do not permit the incorporation of covariates (\eg \cite{zheng1995estimates}, \cite{rivest2001martingale}, \cite{de2013generalized}), or make strict assumptions over the form of the marginal distributions of $f_{T_E}$ and $f_{T_C}$ (\eg \cite{escarela2003fitting}). These limitations mean it is difficult to apply these ideas to survival times modeled via nonlinear functions (such as neural networks) that are increasingly being used. In this vein, our work makes the following contributions:
\begin{enumerate}[wide, labelwidth=!, labelindent=0pt]
    \item We show how to leverage copulas to correct for dependent censorship in neural network based models of survival outcomes. We present a parameteric proportional hazards model that leverages neural networks to relax assumptions on the distributional form of the marginal event and censoring functions, and employs a copulas to model the dependence between event and censoring. We also present a method to jointly learn the model and copula parameter from right-censored survival data. To our knowledge, this work represents the first neural network-based model of survival analysis to account for dependent censoring.
    \item We demonstrate that conventional survival metrics, like concordance, are biased under dependent censoring, and we highlight the general impossibility of unbiased evaluation in this regime.
    \item It is statistically impossible to determine whether $T_E$ and $T_C$ are independent or dependent from data alone. We show how the \textit{choice of copula can represent an assumption} (prescribed via domain knowledge) over the relationship between the event and censoring distributions. Our paper cleanly characterizes the dependence assumptions underlying two common families of copula (\mbox{the Clayton and Frank families}), and provides guidance to practitioners in choosing a copula to meet their needs. The incorporation of the copula enables practitioners to improve the resulting model on a variety of different benchmarks. 
\end{enumerate}

\section{Background and Preliminaries}
For notation, we will use \color{frenchblue}$T_E$ \color{black} and \color{mediumred-violet}$T_C$ \color{black} where appropriate to refer (respectively) to the random variables representing time-of-\color{frenchblue}event \color{black} and \color{mediumred-violet}censorship\color{black}. When a time could refer to either, we will instead simply use $T$. Realizations of each random variable, such as the time-of-event for a specific patient, will be denoted with a superscript (\eg $T_E^{(i)}$).

\subsection{Survival Analysis Preliminaries}
Our work will use  the following elementary quantities defined by survival analysis: $f_{T|X}$, $F_{T|X}$, representing the conditional density and cumulative distribution functions over the time of an outcome of interest (\eg event or censorship). Then, we have the following definitions.

\begin{definition}[Survival Function]
The \textit{survival function}
\begin{equation}
S_{T|X}(t|X)\ \triangleq\ \Cprob{T > t}{X}\ =\ 
1-F_{T|X}(\,t\,|\,X\,)
\end{equation}
represents the likelihood that event (or censorship) will take place after a specified time, $t$.
\end{definition}

\begin{definition}[Hazard Function]
The \textit{hazard function},
\begin{equation}
\small
h_{T|X}(t|X) \triangleq \lim_{\epsilon \rightarrow 0}
  \Cprob{T \in [t, t + \epsilon) }{ T \geq t, X} \ =\ \frac{f_{T|X}(t|X)}{S_{T|X}(t|X)}
 \label{eq:hazard_function}
\end{equation}
represents the probability that the event will take place within an infinitesimal window in the future, given that it has not yet occurred.
\end{definition}

\begin{definition}[Likelihood Function]
The general likelihood function for survival data $\mathcal{D}\,=\,\{ (X^{(i)}, T_\text{obs}^{(i)}, \delta^{(i)}) \}_{i=1}^N$ is the following
\footnote{
{The standard presentation of the survival likelihood is the survival likelihood under conditional independence (Equation \mbox{\ref{eq:independence_likelihood}}), which represents a special case of Equation \mbox{\ref{eq:generallikelihood}}. For a derivation of Equation \mbox{\ref{eq:generallikelihood}}, refer to Appendix \ref{appx:the_right_censored_likelihood}}.}
\begin{align}\footnotesize
\mathcal{L}(\mathcal{D}) = \prod_{i=1}^N &\color{frenchblue}{\underbrace{\color{black}\left[\int_{T^{(i)}_\text{obs}}^\infty f_{T_E, T_C | X}(T^{(i)}_{\text{obs}},\, t_c\, |\, X^{(i)})\,dt_c\right]\color{frenchblue}}_{\Pr\left(T_E = T^{(i)}_{\text{obs}},\, T_C > T^{(i)}_{\text{obs}}\, |\, X^{(i)}\right)}}^{\color{black}\delta^{(i)}}\label{eq:survivallikelihood}\\&\color{mediumred-violet}{\underbrace{\color{black}\left[\int_{T^{(i)}_{\text{obs}}}^\infty f_{T_E, T_C | X}(t_e,\, T^{(i)}_{\text{obs}}\, |\, X^{(i)})\,dt_e\right]}_{\color{mediumred-violet} \Pr\left(T_C = T^{(i)}_{\text{obs}},\, T_E > T^{(i)}_{\text{obs}}\, |\, X^{(i)}\,\right)}}^{\color{black}1-\delta^{(i)}} \nonumber
\end{align}
\label{eq:generallikelihood}
\end{definition}

\subsection{Copulas and Sklar's Theorem}
\begin{definition}[Copula \citep{nelsen2007introduction}]
A copula $C(u_1, ..., u_m) : [0, 1]^m \rightarrow [0, 1]$
is a function with the following properties.
\begin{enumerate}[wide, labelwidth=!, labelindent=0pt]
    \item \ul{Groundedness}: if there exists an $i \in \{1, ..., m\}$ such that $u_i = 0$, then $C(u_1, ..., u_m) = 0$.
    \item \ul{Uniform Margins}: for all $i \in \{1, ..., m\}$, if $\forall j:\ j\neq i \Rightarrow u_{j} = 1$, then $C(u_1, ..., u_m) = u_i$.
    \item \ul{$d$-Increasingness}: for all $u = (u_1, ..., u_m)$, $v = (v_1, ..., v_m)$ where $u_i < v_i$ for all $i = 1, ..., m$, the following holds:
    \begin{equation*}
    \sum_{l \in \{0, 1\}^m} (-1)^{l_1 + ... + l_m} C(u_1^{l_1}v_1^{1-l_1}, ..., u_m^{l_m}v_m^{1-l_m}) \geq 0
    \end{equation*}
\end{enumerate}
\label{def:copula}
\end{definition}

The utility of copulas as probabilistic objects stems primarily from the application Sklar's Theorem \citep{sklar1959fonctions}, which demonstrates that any joint cumulative density can be written in terms of a copula over the quantiles of its marginal cumulative densities.

In this work, we will place our emphasis on those copulas that model \textit{joint survival functions}. Such copulas are known as \textit{survival copulas}, and their own version of Sklar's theorem (Equation \ref{eq:sklar-survival}) applies.

\begin{figure}
\centering
\includegraphics[width=0.47\textwidth]{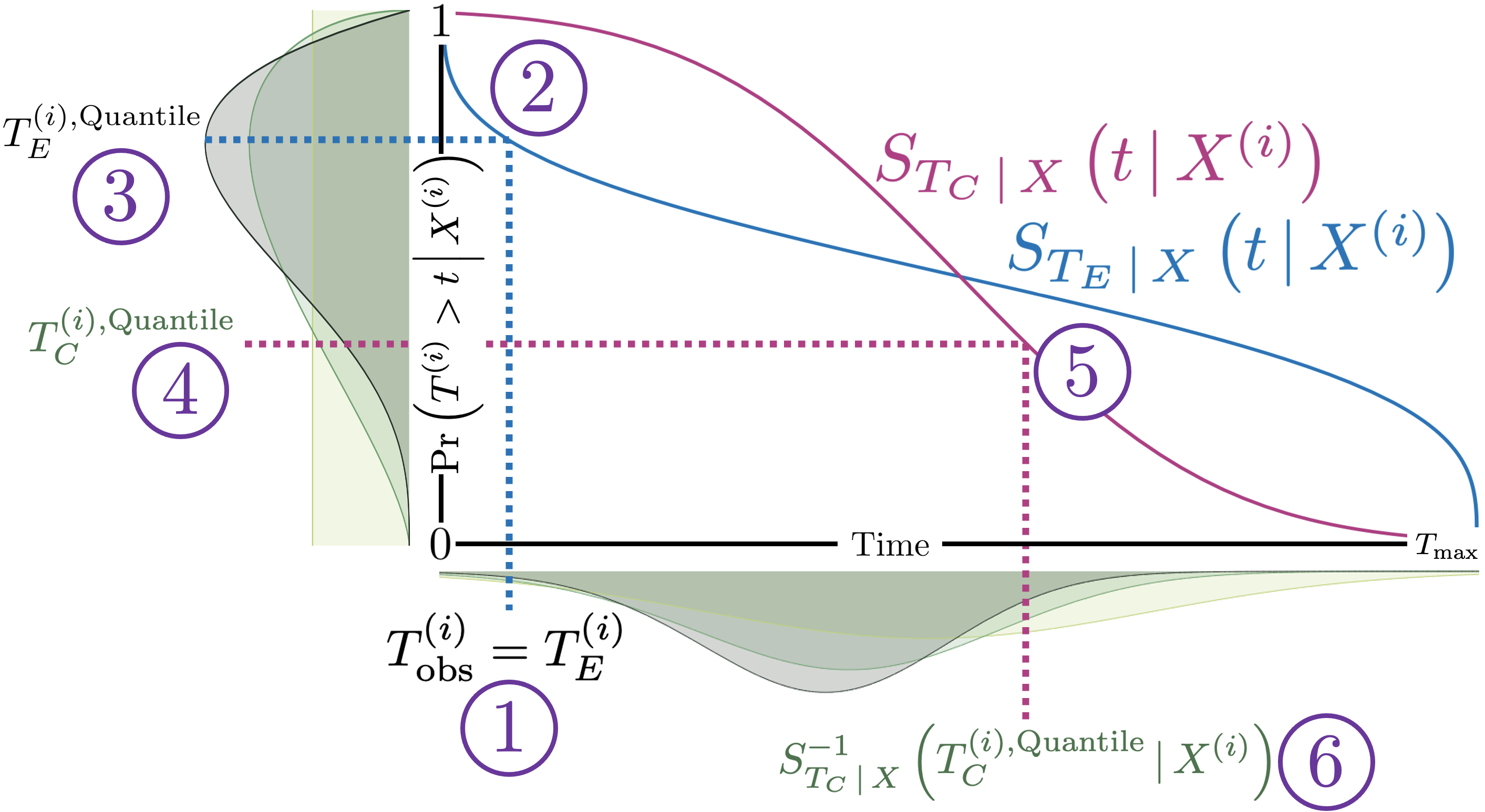}
\caption{\small Visualization of how Sklar's Theorem (Survival) models quantile dependency using a copula. \textcolor{violet}{\textbf{(1)}} The observed event time, $T_E^{(i)}$, is \textcolor{violet}{\textbf{(2)}} mapped through the event survival function, $S_{T_E|X}$, to \textcolor{violet}{\textbf{(3)}} obtain an \textit{event quantile}, $T_E^{(i),\text{Quantile}}$. \textcolor{violet}{\textbf{(4)}} A \textit{censoring quantile} is sampled from the copula, $T_C^{(i),\text{Quantile}} \sim C_\theta(\cdot | T_E^{(i),\text{Quantile}})$; the distributions to the left of the vertical axis show the probability mass of $C_\theta(\cdot | T_E^{(i),\text{Quantile}})$ under \textcolor{independence-text}{\textbf{no}}, \textcolor{weakdependence-text}{\textbf{weak}}, and \textcolor{moderatedependence-text}{\textbf{moderate}} dependence. Notice how as the dependence increases, the distribution $C_\theta(\cdot | T_E^{(i),\text{Quantile}})$ concentrates mass around $T_E^{(i),\text{Quantile}}$. \textcolor{violet}{\textbf{(5)}} The censoring quantile is then mapped through the inverse censoring survival function, $S_{T_C|X}^{-1}$, to \textcolor{violet}{\textbf{(6)}} obtain a corresponding time-of-censorship, $T_C^{(i)}$. The distributions below the horizontal axis show the distribution of $T_C^{(i)}$ under \textcolor{independence-text}{\textbf{no}}, \textcolor{weakdependence-text}{\textbf{weak}}, and \textcolor{moderatedependence-text}{\textbf{moderate}} dependence.
}
\label{fig:quantile-figure}
\end{figure}

\begin{theorem}[Sklar's Theorem (Survival Copulas) \citep{nelsen2007introduction}]
A survival copula\footnote{The copula that relates the joint cumulative distribution $F_{X_1, ..., X_m}$, with the marginal cumulative distribution functions is typically not the same as that which relates the joint survival function $S_{T_1, ..., T_M}$ with the marginal survival functions, though both are valid copulas \citep{nelsen2007introduction}.} is a copula that applies Sklar's Theorem to survival functions, as follows:
\begin{equation}
\label{eq:sklar-survival}\small
S_{T_1, ..., T_m}(t_1,\, \dots\,,\, t_m\,)\ =\ C(\,S_{T_1}(t_1),\, \dots,\, S_{T_m}(t_m)\,)
\end{equation}
\end{theorem}
A visualization of the way in which a copula induces dependency between $T_E$ and $T_C$ via the quantiles of $S_{T_E | X}$ and $S_{T_C | X}$, is shown in Figure \ref{fig:quantile-figure}.

We will focus on two families of copulas, the Clayton \citep{clayton1978model} and Frank \citep{frank1979simultaneous} families. Within these families, the copula $C_\theta$ is parameterized by a single parameter, $\theta$, interpreted as the degree of dependence between the marginal distributions under Equation \ref{eq:sklar-survival}. A larger value of $\theta$ implies greater dependency between the marginal distributions, and both families of copulas converge to the independence copula as $\theta$ approaches 0. We additionally restrict ourselves to \textit{bivariate} survival copulas, although in principle, these methods could be directly extended to accommodate an arbitrary number of competing events. Such uniparametric copulas provide a parameter-efficient means of modeling the joint survival function: given that survival analysis already provides tools to model the marginal survival functions $S_{T_E}$, $S_{T_C}$, a model that couples these distribution functions via a uniparametric copula $C_\theta$ only requires adding one additional parameter to the model.

\section{Related Work}
\label{sec:relatedwork}
\textbf{Deep Learning in Survival Analysis}:
Linear models of survival analysis make the (often unrealistic) assumption that an individual's time-to-event is determined by a linear function of his or her covariates. \cite{faraggi1995neural} presented the first neural-network based model of survival, by incorporating a neural network into a Cox Proportional Hazards (CoxPH) model \citep{cox1972regression}. Although subsequent experimentation found the Farragi-Simon model unable to outperform its linear CoxPH counterpart \citep{mariani1997prognostic, xiang2000comparison}, DeepSurv \citep{katzman2018deepsurv} leveraged modern tools from deep learning such as SELU units \citep{klambauer2017self} and the Adam optimizer \citep{kingma2014adam}, to learn a practical neural network-based CoxPH model that reliably outperformed the linear CoxPH on nonlinear outcome data. Since then, variations of neural network-based models of survival, such as DeepHit \citep{lee2018deephit} (and its extension to time-varying data, Dynamic-DeepHit \citep{lee2019dynamic}), Deep Survival Machines \citep{nagpal2021deep}, SuMo-net \citep{rindt2022survival}, Transformer-based survival models \citep{hu2021transformer, wang2022survtrace}, and methods based off of Neural ODEs \citep{tang2022soden} have been introduced to model survival outcomes. Though these models successfully relax assumptions around the functional form of marginal risk, they do not jointly model the event and censoring times, a limitation that does not allow them to appropriately account for dependent censorship.

DeepSurv has enjoyed enduring success in part due to its broad applicability and strong performance on clinical data (\eg \cite{kim2019deep, hung2019deep, she2020development}). Therefore, our investigation will focus on relaxing the conditional independence assumption in a parameteric proportional hazards model; we leave to future work the relaxation of the conditional independence assumption in other classes of neural network based survival models.

\textbf{Missing/Censored-Not-At-Random Data and Identification}:
Since we do not simultaneously observe $T_E$ and $T_C$, we can treat the problem of survival analysis as one of missing data. The standard taxonomy \citep{rubin1976inference, tsiatis2006semiparametric} of missing data partitions variables into one of three classes: \textit{missing completely at random (MCAR)} where the missingness process is independent of the value of any observed variable, \textit{missing at random (MAR)} where the missingness process may depend on the value of one or more observed covariates, and \textit{missing not at random (MNAR)} where the missingness process may depend on unobserved variables (such as unobserved confounding or self-masking). Similarly, censorship in survival analysis can take place \textit{completely at random (CCAR)}, \textit{at random (CAR)}, or \textit{not at random (CNAR)} \citep{leung1997censoring, lipkovich2016sensitivity}. The conditional independence assumption of Equation \ref{eq:conditional_indep} is equivalent to asserting CAR in the data. 

MNAR data, in the general case, is non-identifiable \citep{nabi2020full}; but survival analysis imposes stronger assumptions on the data than general models of missing data, since observed event time acts as a lower bound for unobserved event time (in the case of censored data). Therefore, prior work has focused on investigating the scenarios in which model parameters of survival data can be uniquely identified. \cite{tsiatis1975nonidentifiability} established that, in the general case, the joint over $M$ variables, $\Pr(T_1, ..., T_M)$ is not generally identifiable from observations of the random variable $T = \min\left(T_1, ..., T_M\right)$; although if the joint distribution is defined in terms of a known copula $C$, and the marginals are continuous, then identifiability holds \citep{zheng1996identifiability, carriere1995removing}. \cite{crowder1991identifiability} extended this line of work and showed that even if all the marginal distributions $f_1, ..., f_M$ are known, the joint distribution remains non-identifiable. Research in statistics has since defined tuples of marginals and copulas for which the joint distribution is identifiable. Notably, \cite{schwarz2013identifiability} and prove that if the marginals $f_E$ and $f_C$ are known, several sub-classes of Archimedean copulas are identifiable in the bivariate case. \cite{zheng1996identifiability, carriere1995removing} highlight conditions for identifiability when the form and parameter of the copula are known \textit{a priori}. \cite{schwarz2013identifiability} categorize copulas into sub-classes wherein the ground-truth copula, $C_{\theta^*}$, is identifiable. Our current analysis does not touch upon the identifiability of the joint distribution in the context of neural network based models of survival outcomes though the success of our method does highlight this as an important area for future study. Many machine learning models remain non-identified \citep{bona2021parameter} while remaining useful as  predictive and descriptive models. We consider our method a similar approach in this respect.

\textbf{Copula-Based Models of Dependent Censoring}:
Prior literature has leveraged copulas to model the relationship between the event and censoring distributions in order to account for the effect of dependent censoring \cite{emura2018analysis}. To our knowledge, the first such work was that of \cite{zheng1995estimates} and \cite{rivest2001martingale}, whose development of the nonparametric Copula-Graphic Estimator extended the Kaplan-Meier Estimator \citep{kaplan1958nonparametric} to cases where the dependence between $T_E$ and $T_C$ takes the form of an assumed copula (both $C$, $\theta$ assumed to be known). Though parametric estimators for this problem have been proposed in prior literature, they tend to make strict assumptions over the distributional form of $f_{T|X}$ (\eg that it is a linear-Weibull function \citep{escarela2003fitting}\footnote{Although Escarela does not directly model dependent censoring, but rather dependent competing events, the approach can be directly extended to this domain.}). Proposed semi-parametric estimators \citep{chen2010semiparametric, emura2017joint, deresa2022copula} suffer from much the same problem, as both of these approaches assume that the hazard is a linear function of the instance covariates. To our knowledge, no such copula-based model exists to accommodate more complex relationships between covariates and risk while also accounting for dependent censoring. This is the gap our research aims to fill.

\section{Model and Optimization}
We now present our extension of the Weibull CoxPH model~\citep{barrett2014weibull}, and discuss the problem of learning nonlinear models of survival outcomes under dependent censorship. Our approach entails modeling each outcome -- event and censorship -- independently with an extension of the Weibull CoxPH model, and linking them via a copula in the likelihood function during training. Our approach makes the following assumptions.

\begin{assumption}[Known Form of the Copula]
\label{assmp:knownform}
We assume prior knowledge of the functional form of the copula (\eg that $C_{\theta^*}$, the copula associated with the data-generating process, is a Clayton copula).\footnote{In some experiments, we weaken this assumption, and we will explicitly note where this is the case.}
\end{assumption}

\begin{assumption}[Proportional Hazards \citep{cox1972regression}]
\label{assmp:proportionalhazards}
The hazard for each outcome (event/censorship) can be decomposed into some \textit{baseline hazard} $\lambda_0$, dependent only on time, and some \textit{covariate hazard} $g$, dependent only on the covariates $X$. That is, there exists some appropriate $\lambda_0, g$ for which $h_{T|X}(t| X) = \ \lambda_0(t) \,\exp(\,g(X)\,)$.
\end{assumption}

\subsection{The Weibull CoxPH Model}
\label{sec:Weibull-Cox}

Let $\lambda_0(t) = \left(\frac{\nu}{\rho}\right)\left(\frac{t}{\rho}\right)^{\nu-1}$ denote the baseline hazard of the Weibull CoxPH model, and let $g_{\psi}$ denote a neural network with parameters $\psi$ mapping the covariate space $\mathcal{X}$ to the real line. Then, leveraging the proportional hazards assumption, we define our model in terms of its hazard:
\begin{equation}
\label{eq:model-hazard}
\hat{h}_{T|X}(\,t|X\,)\ =\ \left(\frac{\nu}{\rho}\right)\left(\frac{t}{\rho}\right)^{\nu-1} \exp\left(\,g_\psi(X)\,\right)
\end{equation}

Let $\phi = \{\nu, \rho, \psi\}$ denote the complete set of model parameters, and observe that the Weibull CoxPH model is fully parametric model over these \textit{marginal parameters} $\phi$. By rearranging Equation \ref{eq:model-hazard}, this class of models readily admits $\hat{S}_{T|X}$, the estimated survival function over each outcome, and $\hat{f}_{T|X}$, the corresponding probability mass function. These two quantities will allow us to perform maximum likelihood estimation -- their derivations are provided in Appendix \ref{appx:the_survival_function} and \ref{appx:the_density_function}.
\begin{align}
\hat{S}_{\,T|X}(\,t|X\,)\ &=\ \exp\left(-\left(\frac{t}{\rho}\right)^\nu g_\psi(X)\right)\\
\hat{f}_{T|X}(\,t|X\,)\ &=\ h_{T|X}(\,t|X\,)\ \hat{S}_{T|X}(\,t|X\,)
\end{align}

\subsection{Maximum Likelihood Learning Under Dependent Censorship}
\label{sec:likelihood}

Let $\mathcal{D} = \{(X^{(i)}, T_{\text{obs}}^{(i)}, \delta^{(i)})\}_{i=1}^N$ represent a dataset comprising $N$ i.i.d. draws from some data-generating distribution. Let $X^{(i)} \in \mathcal{X}$ refer to a set of baseline covariates collected about each individual $i$. Let $T_{\text{obs}}^{(i)} \in \mathbb{R}_+$ refer to their time of last observation, taken to be the minimum of latent variables $T_E^{(i)} \in \mathbb{R}_+$, $T_C^{(i)} \in \mathbb{R}_+$, representing the event and censoring times, respectively. Finally, let $\delta^{(i)} \in \{0, 1\}$ represent an event indicator taking on the value $\mathbbm{1}[T_E^{(i)} < T_C^{(i)}]$. Let $C$ represent a survival copula. Given $\mathcal{D}$, we learn by maximizing the likelihood of the observed data.

Under conditional independence, Equation \ref{eq:survivallikelihood} factorizes and simplifies into the familiar form of the survival likelihood.
\footnotesize
\begin{align}
\mathcal{L}(\mathcal{D}) = \prod_{i=1}^N &\left[f_{T_E|X}(T^{(i)}_{\text{obs}} | X^{(i)})S_{T_C|X}(T^{(i)}_{\text{obs}}|X^{(i)})\right]^{\delta^{(i)}} \label{eq:independence_likelihood}\\& \left[f_{T_C|X}(T^{(i)}_{\text{obs}} | X^{(i)})S_{T_E|X}(T^{(i)}_{\text{obs}}|X^{(i)})\right]^{1-\delta^{(i)}}\nonumber
\end{align}
\normalsize

However, when $T_E,T_C$ are no longer conditionally independent, we can no longer rely on this clean decomposition of the log-likelihood. Instead, we make use of the following lemma.
\begin{lemma}[Conditional Survival Function Under Sklar's Theorem (Survival)]
\label{lemma:copula-conditional}
If $S_{T_E, T_C | X}(t_e, t_c | x) = \left.C(u_1, u_2)\middle|_{\substack{{u_1=S_{T_E|X}(t_e|x)}\\ {u_2=S_{T_C|X}(t_c|x)}}}\right.$, then,
\footnotesize
\begin{equation}
\int_{t_c}^\infty f_{T_C | T_E, X}(t_c | t_e, x) = \left.\frac{\partial}{\partial u_1} C(u_1, u_2)\middle|_{\substack{{u_1=S_{T_E|X}(t_e|x)}\\ {u_2=S_{T_C|X}(t_c|x)}}}\right..
\nonumber
\end{equation}
\end{lemma}
\normalsize

Applying Lemma~\ref{lemma:copula-conditional} to Equation \ref{eq:survivallikelihood} yields the log-likelihood for survival models under dependent censorship.\pagebreak
\footnotesize
\begin{align}
\label{eq:loglikelihood}
\ell(\mathcal{D}) &= \sum_{i=1}^N 
{ } \delta^{(i)}\log \left[f_{T_E|X}(T^{(i)}_\text{obs}| X^{(i)})\right] + \\&\delta^{(i)} \log \left[\frac{\partial}{\partial u_1}C(u_1, u_2)\middle\vert_{\substack{u_1 = S_{T_E|X}(T^{(i)}_\text{obs}|X^{(i)})\\u_2 = S_{T_C|X}(T^{(i)}_\text{obs})|X^{(i)}} }\right] +\nonumber\\
&(1-\delta^{(i)})\log \left[f_{T_C|X}(T^{(i)}_\text{obs}| X^{(i)})\right] + \nonumber\\
&(1-\delta^{(i)}) \log \left[\frac{\partial}{\partial u_2}C(u_1, u_2)\middle\vert_{\substack{u_1 = S_{T_E|X}(T^{(i)}_\text{obs}|X^{(i)})\\u_2 = S_{T_C|X}(T^{(i)}_\text{obs})|X^{(i)}} }\right].\nonumber
\end{align}
\normalsize

In this expression, the first term corresponds to the log likelihood of observing the event at time $T_{\text{obs}}^{(i)}$. The second term corresponds to the conditional probability of observing the censorship time after the event time, given that the event time is $T_{\text{obs}}^{(i)}$. The third and fourth terms, by symmetry, represent the same quantities for the censorship time. Despite the visual complexity of Equation \ref{eq:loglikelihood}, the partial derivatives of the Clayton and Frank copulas admit closed form solutions, so the log likelihood function has a closed form and can be maximized via gradient-based methods. Algorithm~\ref{alg:optimization} details the optimization procedure used to jointly optimize the marginal and copula parameters. Empirically, we find that scaling the gradient of $\hat{\theta}$ by a large constant factor $K$, and then clipping it prior to taking each update step, supports stable optimization in this regime ($K = 1000$ in our experiments). Additional implementation details and hyperparameters are discussed in Appendix \ref{appx:implementation_details}.

\RestyleAlgo{ruled}
\SetKwComment{Comment}{$\ $\# }{ }
\SetKwComment{Commentt}{\# }{ }
{
\begin{algorithm}[h]
\small
\KwIn{
$\mathcal{D}$: survival dataset of the form $\{(X^{(i)}, T_\text{obs}^{(i)}, \delta^{(i)})\}_{i=1}^N$; $C_\theta$: a bivariate copula, parameterized by $\theta$; $\mathcal{M}$, a class of survival model parameterized by $\phi$ that can produce $\hat{S}^{(\mathcal{M})}_{T|X}(t|X)$, $\hat{f}^{(\mathcal{M})}_{T|X}(t|X)$, for each $X^{(i)} \in \mathcal{D}$; $\alpha$: learning rate for event model, censoring model, and copula parameter; $M$: number of training epochs; $K$: large constant factor; $\theta_{\text{min}}$: small positive number.
}
\KwResult{$\hat{\theta}, \hat{\phi}_E, \hat{\phi}_C$: learned parameters of the copula and each marginal survival model.}
\hrulefill\\
$\mathcal{M}_E \gets \texttt{Instantiate}(\mathcal{M}; \hat{\psi}_E^{(0)})$ \;
$\mathcal{M}_C \gets \texttt{Instantiate}(\mathcal{M}; \hat{\psi}_C^{(0)})$ \;
$C_\theta \gets \texttt{Instantiate}(C; \hat{\theta}^{(0)})$ \;
\For{$i = 1,\, ...\,,\, M$}{
    $\mathcal{L}_{i} \gets \ell\left[\mathcal{D}; \hat{f}^{\left(\mathcal{M}_E\right)}_{T|X},  \hat{f}^{\left(\mathcal{M}_C\right)}_{T|X}, \hat{S}^{\left(\mathcal{M}_E\right)}_{T|X},   \hat{S}^{\left(\mathcal{M}_C\right)}_{T|X},  
    C_{\hat{\theta}^{(i)}}\right]$\;
    $\hat{\psi}_C^{(i)} \gets $\texttt{AdamUpdate}($\mathcal{L}_i$, $\hat{\psi}_C$, $\alpha$) \;
    $\hat{\psi}_E^{(i)} \gets $\texttt{AdamUpdate}($\mathcal{L}_i$, $\hat{\psi}_E$, $\alpha$) \;
    $\nabla \hat{\theta}^{(i)} \gets \nabla \hat{\theta}^{(i)} \times K$\;
    $\nabla \hat{\theta}^{(i)} \gets \nabla \hat{\theta}^{(i)} \vert_{[-0.1, 0.1]}$\;
    $\hat{\theta}^{(i)} \gets $\texttt{AdamUpdate}($\mathcal{L}_i$, $\hat{\theta}$, $\alpha$) \;
    $\hat{\theta}^{(i)} \gets \min(\hat{\theta}^{(i)}, \theta_{\text{min}})$ \Comment{Constrain theta > 0}
}
\Return{$\hat{\theta}^{(i)}$, $\hat{\psi}_E^{(i)}$, $\hat{\psi}_C^{(i)}$}
\caption{\small Learning Under Dependent Censorship \normalsize}
\label{alg:optimization}
\normalsize
\end{algorithm}
}

\section{Evaluation}

\subsection{Metrics are Biased Under Dependence}
Standard metrics such as the concordance index \citep{harrell1982evaluating, uno2011c}, time-dependent concordance (TDCI) \citep{gerds2013estimating}, and Brier score \citep{brier1950verification} cannot  effectively evaluate models learned under dependent censoring. To demonstrate this, we generate survival data under a copula, and compare the performance of the data-generating event model, $f_{T_E|X}$,  on censored and uncensored data as the dependency increases. The results of this experiment are shown in Table \mbox{\ref{tab:biased_metrics}}. As the dependence increases, both the concordance and Brier score under censoring deviate from their values without censoring. This suggests that the utility of these metrics decreases as the dependence in censoring increases. This challenges previous results that use these measures as the primary statistics of interest when assessing the performance of models under dependent censoring.

By way of analogy, we describe the connection between evaluation under dependent censoring and the potential outcomes framework from causal inference. In the case where censoring takes place completely at random, metrics like concordance and Brier score are suitable means of evaluation, akin to how a randomized controlled trial produces an unbiased estimate of the average treatment effect. Under observed confounding, weighting schemes like inverse-propensity censorship weighting \mbox{\cite{uno2011c, graf1999assessment}} leverage a censoring model to produce an unbiased estimator of the evaluation statistic. But confounding of the form in survival analysis does not readily admit a censoring model that can be used to perform weighting adjustment since the covariates required for such a model remain unobserved. Consequently, unbiased model evaluation under dependent censoring is fundamentally a problem of counterfactual analysis and not feasible to solve using observational data alone.

\begin{table}[]
    \centering\scriptsize
    \setlength\tabcolsep{2pt}
    \begin{tabular}{|c||c|c|c|c|c|c|}
    \hline
    & \multicolumn{3}{c|}{C-Index ($\uparrow$)} & \multicolumn{3}{c|}{Brier Score ($\downarrow$)}\\
    $\tau$ & \multicolumn{1}{c}{Uncensored} & \multicolumn{1}{c}{Censored} & \multicolumn{1}{c|}{Abs. Diff. ($\downarrow$)} & \multicolumn{1}{c}{Uncensored} & \multicolumn{1}{c}{Censored} & \multicolumn{1}{c|}{Abs. Diff. ($\downarrow$)}\\
    \hline\hline
    0.01 & 0.6151 & 0.6187 & 0.0037 & 0.0719 & 0.0859 & 0.0140\\
    0.2 & 0.6144 & 0.6140 & 0.0004 & 0.0757 & 0.0909 & 0.0152\\
    0.4 & 0.6170 & 0.6164 & 0.0006 & 0.0726 & 0.0943 & 0.0217 \\
    0.6 & 0.6172 & 0.6342 & 0.0170 & 0.0733 & 0.0963 & 0.0230\\
    0.8 & 0.6125 & 0.6873 & 0.0748 & 0.0744 & 0.1054 & 0.0310\\
    \hline
    \end{tabular}
    \caption{\small The results of an experiment comparing the concordance index and Brier score on an uncensored population, against that on a population experiencing dependent censoring. The full details of this experiment are provided in Appendix \ref{appx:evaluation_metrics_are_biased_under_dependence}}
    \label{tab:biased_metrics}
\end{table}

\begin{figure}
    \centering
    \includegraphics[width=0.4\textwidth]{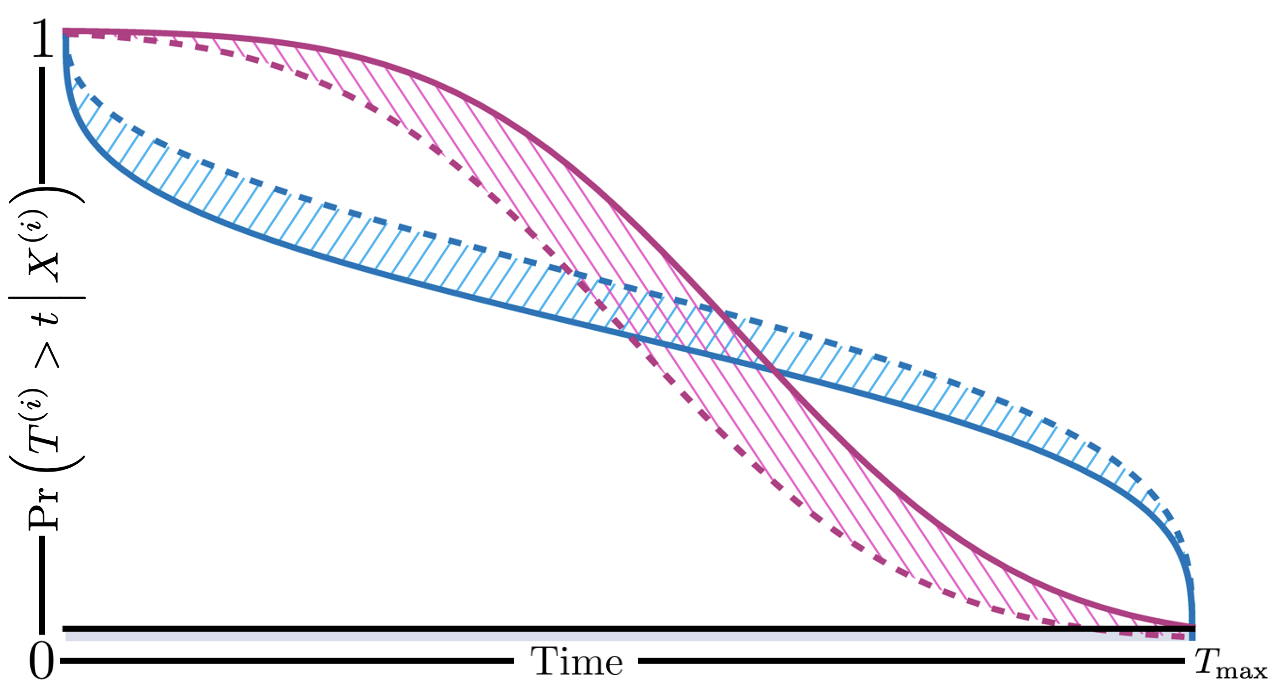}
    \caption{\small The \textit{Survival-$\ell_1$} metric, $\mathcal{C}_{\text{Survival-}\ell_1}(S, \hat{S})$, for \textcolor{frenchblue}{event} and \textcolor{mediumred-violet}{censoring} distributions. Dashed lines represent the predicted survival curves, \textcolor{frenchblue}{$\hat{S}_{T_E|X}$}, and \textcolor{mediumred-violet}{$\hat{S}_{T_C|X}$}, while solid lines represent the corresponding ground-truth survival curves, \textcolor{frenchblue}{$S_{T_E|X}$}, and \textcolor{mediumred-violet}{$S_{T_C|X}$}. The black horizontal line represents the normalizing quantile, $Q_{\lVert\cdot\rVert}$, which is used to standardize the duration of the survival curve across patients when calculating the \textit{Survival-}$\ell_1$. The area of the hatched blue region above $Q_{\lVert\cdot\rVert}$ is the value of \textcolor{frenchblue}{$\mathcal{C}_{\text{Survival-}\ell_1}(S_{T_E|X}, \hat{S}_{T_E|X})$}, while that of the hatched pink region is the value of \textcolor{mediumred-violet}{$\mathcal{C}_{\text{Survival-}\ell_1}(S_{T_C|X}, \hat{S}_{T_C|X})$}.}
    \label{fig:eval-metric}
\end{figure}

\textbf{The \textit{Survival-}$\ell_1$ Metric}: We introduce the \textit{Survival-$\ell_1$} as a means of quantifying bias in survival analysis due to dependent censoring on synthetic data. The \textit{Survival-$\ell_1$} metric $\mathcal{C}_{\text{Survival-}\ell_1} : \mathcal{S} \times \mathcal{S} \rightarrow \mathbb{R}_+$, is the $\ell_1$ distance between the ground-truth survival curve, $S_{T|X}$, and the estimate achieved by a survival model, $\hat{S}_{T|X}$ (Figure \ref{fig:eval-metric}), over the lifespan of the curves.

However, the scale of the naive $\ell_1$ measure between survival curves is proportional to the total amount of elapsed time under each survival curve. To ensure that survival curves over longer lifespans do not contribute proportionally more to the evaluation metric than those over shorter lifespans, we define the small constant \textit{normalizing quantile}, $Q_{\lVert\cdot\rVert}$ (in our experiments, $Q_{\lVert\cdot\rVert} = 0.01$). We can loosely think of the time when each survival curve reaches the normalizing quantile as the ``end time'' of that survival curve. By normalizing the area between the survival curves by the \textit{temporal normalization} value $T_{\text{max}}^{(i)} = S^{-1}_{T|X^{(i)}}\left(Q_{\lVert\cdot\rVert}\right)$, we ensure that the duration spanned by a patient's survival curve does not influence that patient's contribution to $\mathcal{C}_{\text{Survival-}\ell_1}$ relative to other patients.

Our \textit{Survival-$\ell_1$} metric therefore takes the following form:
\footnotesize
\begin{align}
\mathcal{C}_{\text{Survival-}\ell_1}(S, \hat{S}) = \sum_{i=1}^N &\frac{1}{N \times T_{\text{max}}^{(i)}} \int_{0}^{\infty} \\&\left|S_{T|X}(t|X^{(i)}) - \hat{S}_{T|X}(t|X^{(i)})\right| dt\nonumber
\end{align}
\normalsize

\begin{figure*}[h!]
    \centering
    \includegraphics[width=\textwidth]{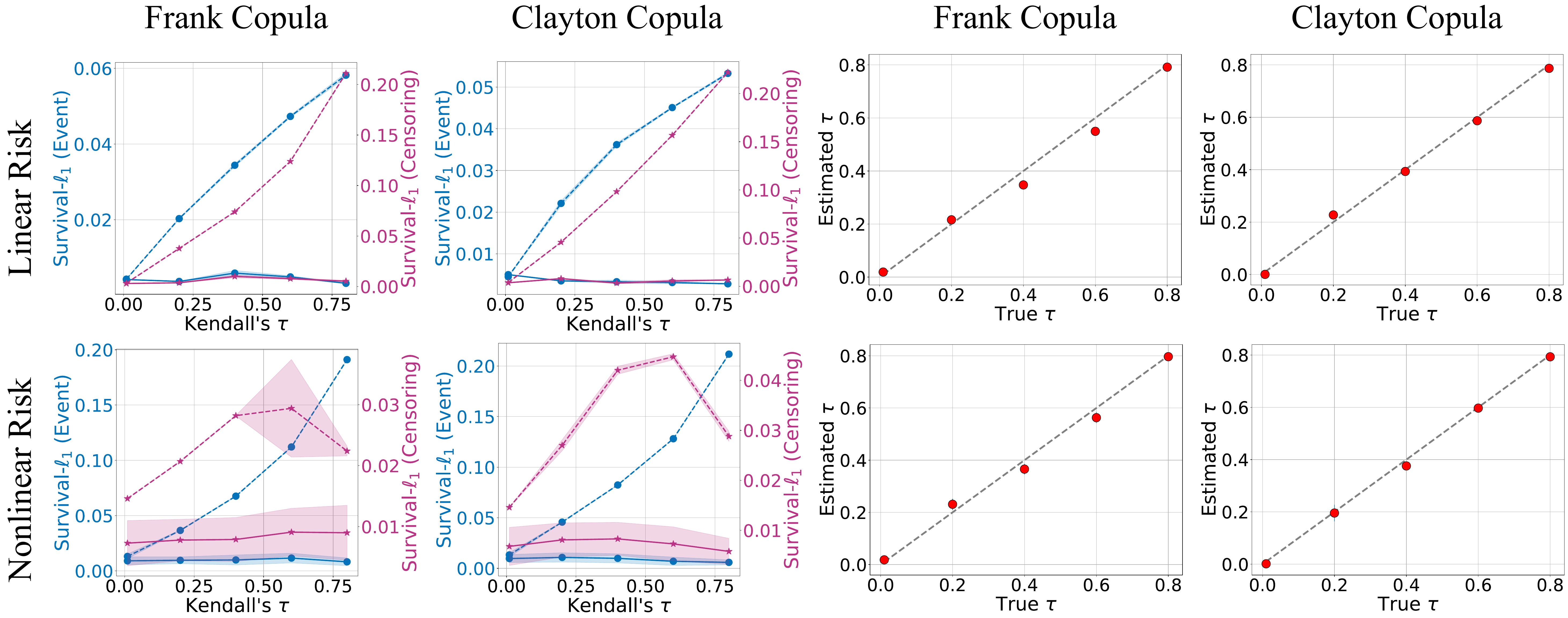}
    \caption{\small \textbf{Left}: Plot of the bias, $\mathcal{C}_{\text{Survival-}\ell_1}$, as a function of the dependence (Kendall's $\tau$), for both independence-assuming and copula-based models on synthetic data. 
    Going from left to right on the $x$-axis denotes stronger dependence between the survival and event time in the data generating process. The $y$-axis is overloaded; the scales on the left hand side of each $y$-axis correspond to bias incurred in the prediction of the event times and the scales on the right hand side correspond to bias incurred in the prediction of the censoring times. Dotted lines represent the bias in the \textcolor{frenchblue}{event} and \textcolor{mediumred-violet}{censoring} survival curves incurred by independence-assuming models, while solid lines represent the bias incurred by our copula-based approach. The copula-based approach yields a lower line for each event, indicating a better approximation of the ground-truth survival function.The shaded region represents the standard deviation of the \textit{Survival-}$\ell_1$ across 10 instantiations of the model with different random seeds. \textbf{Right}: For each value of $\tau$ in the left plot, we plot the recovered value of Kendall's $\tau$, $\hat{\tau}$, as a function of the true dependence, $\tau^*$. The dashed diagonals line, representing $\hat{\tau} = \tau^*$, is plotted for reference. Points close to the line indicate that the learned dependence parameter was close to that of the data-generating process.}
    \label{fig:mainresults}
\end{figure*}

\section{Experiments and Results}
\textbf{Synthetic Data}:
The \textit{Survival}-$\ell_1$ metric places strong assumptions on our knowledge of the data-generating process by assuming access to the ground-truth survival functions for each outcome. For this reason, we predominantly make use of synthetic data to evaluate the merits of our approach.

Algorithm \ref{alg:datagenerating} provides a means of generating synthetic data under a specified copula $C$ with Weibull CoxPH margins. For the \texttt{Linear-Risk} experiment shown in Figure \ref{fig:mainresults}, we generate data according Algorithm \ref{alg:datagenerating} with $X \in \mathbb{R}^{N \times 10} \sim \mathcal{U}_{[0,1]}$, $\nu_E^* = 4, \rho_E^* = 14, \psi_E^*(X) = \beta_E^T(X)$, $\nu_C^* = 3, \rho_C^* = 16, \psi_C^*(X) = \beta_C^T(X)$, where $\beta_E, \beta_C \in [0,1]^{10} \sim \mathcal{U}_{[0,1]}$. For the \texttt{Nonlinear-Risk} experiment, we run Algorithm \ref{alg:datagenerating} with $X \in \mathbb{R}^{N \times 10} \sim \mathcal{U}_{[0,1]}$, $\nu_E^* = 4, \rho_E^* = 17, \psi_E^*(X) = \sum_{i=1}^{10}X_{i}^{2}/8$, $\nu_C^* = 3, \rho_C^* = 16, \psi_C^*(X) = \beta_{C}^{T}X^{2}/5$, where $ \beta_C \in [0,1]^{10} \sim \mathcal{U}_{[0,1]}$. Each synthetic experiment was performed on $20,000$ train, $10,000$ validation, and $10,000$ test samples.

The network $g_\psi$ in the model we train on the \texttt{Linear-Risk} data consists of a single linear layer, while the network $g_\psi$ in the model we train on the \texttt{Nonlinear-Risk} data consists of a three-layer fully-connected neural network with ELU activations and hidden layers consisting of $[10, 4, 4, 4, 2, 1]$ dimensions, respectively.

\textbf{Semi-Synthetic Data}: To investigate the promise of our approach on non-synthetic data, we artificially censor regression datasets according to a various degrees of dependence. We choose two datasets (\texttt{STEEL}) \citep{asuncion2007uci} and \texttt{AIRFOIL} \citep{misc_airfoil_self-noise_291} from the UCI Machine Learning Repository. We induce censoring in the data according to Algorithm \ref{alg:semi-synthetic} in Appendix \ref{appx:creating_a_semi_synthetic_dataset_with_dependent_censoring}. We then train a linear version of our method on the artificially censored dataset and evaluate our performance via the $R^2$ statistic\footnote{Note that a method like \textit{Survival-}$\ell_1$ does not apply to this context, as semi-synthetic data does not provide ground-truth survival curves.}. In this experiment, we compare our approach against two baselines: a linear Weibull CoxPH model trained on the regression data \textit{without censoring}, and a linear independence-assuming Weibull CoxPH model.

\RestyleAlgo{ruled}
\SetKwComment{Comment}{$\ $\# }{ }
\SetKwComment{Commentt}{\# }{ }
{
\begin{algorithm}[h]
\small
\KwIn{
$X \in \mathbb{R}^{N \times d}$: a set of covariates, $g_{\psi} : \mathbb{R}^{N \times d} \rightarrow \mathbb{R}$: a class of risk function parameterized by $\psi$, $C_\theta$: a class of copula parameterized by $\theta$, $(\nu^*_E, \rho^*_E, \psi^*_E), (\nu^*_C, \rho^*_C, \psi^*_C), \theta^*$: data-generating parameters associated with each outcome model and the copula, respectively.}
\KwResult{$\mathcal{D}$, a survival dataset with the desired dependence.}
\hrulefill\\
$\mathcal{D} = \emptyset$\;
\For{$i = 1,\, ...\,,\, N$}{
    $u_1^{(i)}, u_2^{(i)} \sim C_{\theta^*}$\;
    $T_E^{(i)} \gets \left(\frac{-\log(u_1)}{g_{\psi_E^*}(X^{(i)})}\right)^{\frac{1}{\nu^*_E}}\rho^*_E$\;
    $T_C^{(i)} \gets \left(\frac{-\log(u_2)}{g_{\psi_C^*}(X^{(i)})}\right)^{\frac{1}{\nu^*_C}}\rho^*_C$\;
    $\mathcal{D} \gets \mathcal{D} \cup \{(X^{(i)}, \min(T_E^{(i)}, T_C^{(i)}), \mathbbm{1}[T_E^{(i)} < T_C^{(i)}])\}$\;
}
\Return{$\mathcal{D}$}
\caption{\small Generating Synthetic Dependent Survival Data \normalsize}
\label{alg:datagenerating}
\normalsize
\end{algorithm}
}

Our results highlight three properties of our framework. First, our model is capable of reducing the bias in the learned individual survival curve (as measured by the \textit{Survival-$\ell_1$} metric). Second, the learning algorithm does, in many cases, recover the ground truth coefficient associated with the copula when parameterizing the prediction of the event and censoring time with neural networks. Finally, our framework opens up new avenues to learning more complex forms of dependence between event and survival time.

\textbf{Reducing Bias in Survival Outcomes}:
Figure \ref{fig:mainresults} (left column) plots the model bias as measured by the $\textit{Survival-}\ell_1$, and how it behaves across datasets (in rows of plots).

We highlight that our approach of modeling the dependence structure between event and censorship times reduces the bias in the model's estimation of survival curves. The bias is substantially lower under our approach for all values of $\tau > 0$, and we note that the improvements are more pronounced for larger values of $\tau$ indicating that the improvements in our approach are larger as the dependence between censorship and event time is stronger. We see consistent results holding for both the \texttt{Linear-Risk} and \texttt{Nonlinear-Risk} data-generating processes, and for both the Frank and Clayton families of copula. In the special case where $\tau = 0$, we observe that our approach correctly recovers the independence copula, and learns an unbiased survival curve.

Our results on the artificially censored \texttt{STEEL} and \texttt{AIRFOIL} datasets suggest that our method also shows promise on non-synthetic data. On the \texttt{STEEL} dataset, our method achieves an $R^2$ of $0.508$ under high dependence ($\tau=0.8$), compared to the $R^2$ of $0.341$ achieved by the independence-assuming model. Likewise, on the \texttt{AIRFOIL} dataset, our method achieves an $R^2$ of $0.484$ under high dependence ($\tau=0.8$), compared to the $R^2$ of $0.330$ achieved by the independence-assuming model. Across different degrees of dependence, our approach reliably outperforms the independence-assuming baseline, and often approaches the performance of the model trained on the uncensored version of the data. The complete table of results can be found in Appendix \ref{appx:additional_semi_synthetic_results_steel_dataset} (\texttt{STEEL}) and \ref{appx:additional_semi_synthetic_results_airfoil_dataset} (\texttt{AIRFOIL}).

\begin{figure}
    \centering
    \includegraphics[width=0.4\textwidth, keepaspectratio]{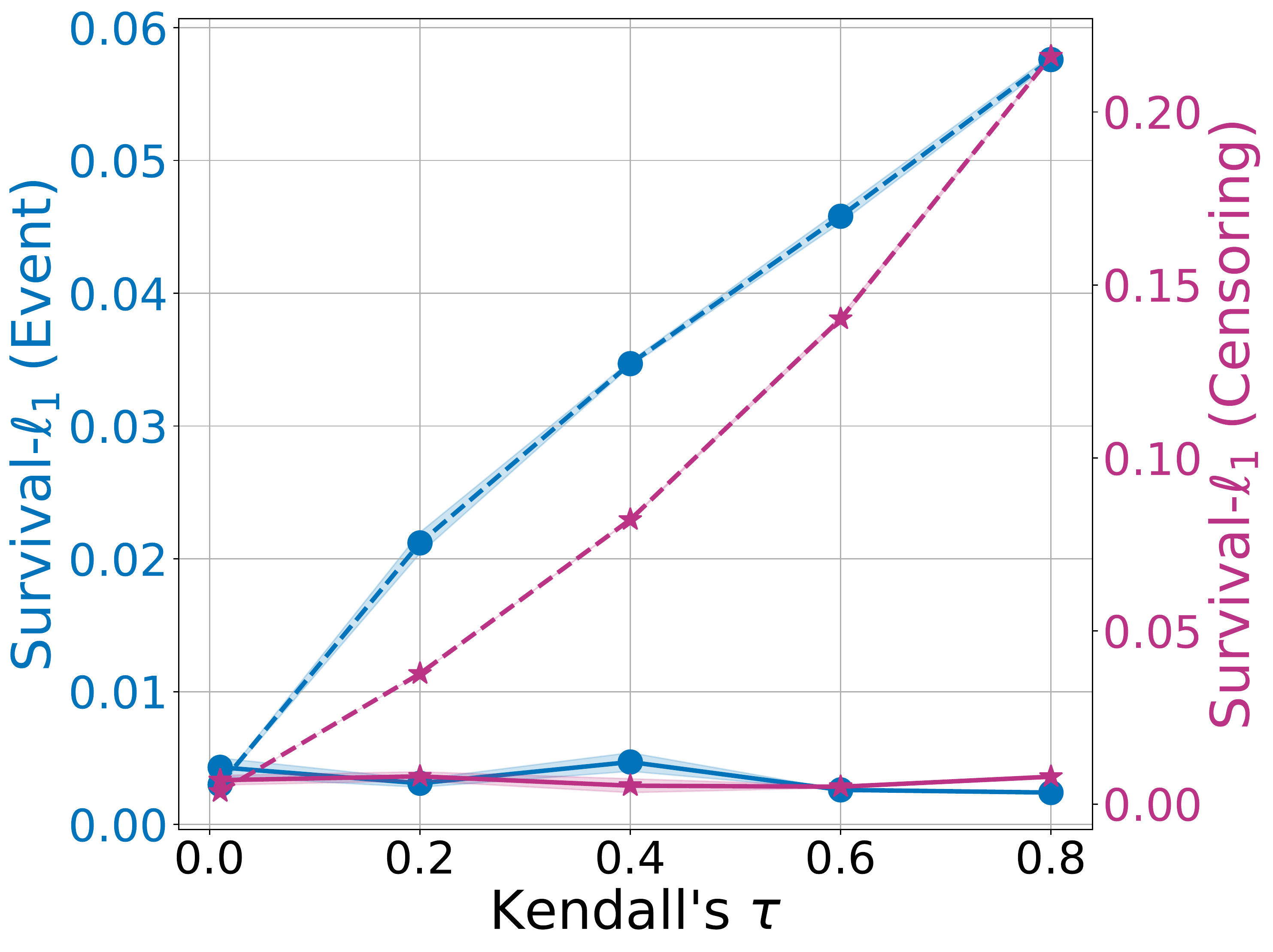}
    \caption{\small Plot of the bias ($\mathcal{C}_{\text{Survival-}\ell_1}$) as a function of dependence (Kendall's $\tau$), for independence-assuming and copula-based Weibull CoxPH models on synthetic data with linear margins drawn from a convex combination of copulas. In this experiment, we optimize over a mixture of two copulas (one Frank, one Clayton), rather than a single uniparametric copula. As in Figure 4, the dashed lines represent the bias incurred by independence-assuming models, while the solid lines represent the bias incurred by our approach. This figure highlights that our method is capable of relaxing Assumption \ref{assmp:knownform} by way of a convex combination of copulas.}
    \label{fig:convex-combination}
\end{figure}

\textbf{Empirical Recovery of the Copula Parameter}:
How close are the recovered parameters of the copula to the true parameters used in the data-generating process? Although we do not have a formal proof of identifiability, we nevertheless study this question empirically on the two datasets in Figure \ref{fig:mainresults} (right column). Here, we find that our approach is able to reliably recover a $\hat{\theta}$ that is close to $\theta^*$ across different datasets and families of copula.

\textbf{Relaxating Assumption \ref{assmp:knownform}}:
Next, we showcase the flexibility of our framework via a relaxation of Assumption \ref{assmp:knownform}. Specifically, rather than parameterizing our model with $C_\theta$, a single copula of an assumed functional form, we instead parameterize it with a convex combination of Clayton and Frank copulas. During optimization, we learn $\theta_{\text{Frank}}$, $\theta_{\text{Clayton}}$, and $\kappa$, a mixing parameter.
Because the Clayton and Frank copulas are both Archimedean, we know that their convex mixture is also a valid Archimedean copula \citep{bacigal2010some, bacigal2015generators}. Figure \ref{fig:convex-combination} shows the results of an experiment on synthetic data with \texttt{Linear-Risk} margins and a dependency produced by a convex combination of copulas: $C_{\text{Mix.}}(u,v) = \kappa C_{\text{Frank}}(u,v) + (1-\kappa) C_{\text{Clayton}}(u,v)$. In this experiment, we fix $\kappa = 0.5$. As in the case where the functional form of $C$ was known, the mixture model reduces bias in estimation of the event and censoring distributions.

\section{Discussion}
\label{sec:discussion}

\subsection{Dependent Censoring in Practice}
\textbf{Evaluating Survival Models on Observational Data}: Given the impossibility of evaluation from observational data alone, how should a practitioner apply our method? We propose that practitioners adopt simulation -- the present gold standard of evaluation from the causal inference literature -- as a primary means to test the performance of survival models under dependent censoring. Such methods as \mbox{\cite{parikh2022validating} and \cite{mahajan2022empirical}} present means of generating counterfactual synthetic data that is similar to the available observational data. Then, evaluating model performance on the simulated data using counterfactual metrics (like Survival-$\ell_1$) is treated as a viable proxy of model performance on the downstream data.

\textbf{The Assumptions Encoded by the Clayton and Frank Copulas}: Given that we only observe either the time of event or censorship, identifying the joint distribution between these variables is generally not possible. Therefore, the choice of copula represents a \textit{assumption} over the data. How can a practitioner leverage domain knowledge in order to select the right copula to use within our framework? Consider how the copula parameter, $\theta$, relates the event and censoring curves under three different circumstances. (1) If the censoring and event curves are identical, then $\theta$ grows with the probability that the time of event and censorship are the same. (2) If the censoring curve decays faster than the survival curve, $\theta$ grows with the probability that the time of censorship precedes the time of event. (3) If the survival curve decays faster than the censoring curve, $\theta$ grows with the probability that the time of event precedes the time of censorship. For a fixed $\theta$, the Clayton copula expresses this dependency as stronger at later times (lower quantiles), and weaker at earlier times (higher quantiles). The Frank copula expresses strength of the dependency at more uniform strength across all time periods. A visualization of these cases, and of the quantile densities expressed by the Clayton and Frank copulas, can be found in Appendix \ref{appx:quantile_density_visualizations} and \ref{appx:intuition_for_copula_selection}.

\section{Conclusion}
The method of using copulas to couple marginal survival distributions is a general one. As future work, we consider extending this approach to other classes of neural survival models, such as those that do not assume either proportional hazards or a Weibull baseline hazard. Though the \textit{Survival-$\ell_1$} metric is a sufficient metric to demonstrate the promise of our approach, it relies on knowledge of the complete survival curve for each instance; this is typically not available in real-world data. The careful study of the behaviour of conventional evaluation metrics under dependence, and the design of strategies to more faithfully ascertain the performance of a model from observational data alone remain open avenues for future work.

Modern statistical methods in survival analysis increasingly rely on complex, nonlinear functions of risk; however, existing applications of deep learning to survival analysis do not accommodate dependent censoring that may be present in the data. This work relaxes this key assumption, and presents the first neural network-based model of survival to accommodate dependent censoring. 
Our experimental results demonstrate the promise of our method: our approach significantly reduces the \textit{Survival-$\ell_1$} (bias) in estimation and our optimization technique is reliably able to recover the underlying dependence parameter in survival data across datasets of varying feature sizes.

\addtocontents{toc}{\protect\setcounter{tocdepth}{3}}

\nocite{seabold2010statsmodels}
\nocite{asuncion2007uci}
\nocite{Dua:2019}
\nocite{ve2021efficient}
\nocite{sathishkumar2020energy}
\nocite{sathishkumar2020industry}
\clearpage

\begin{acknowledgements}
The authors gratefully acknowledge the support of several funding sources without which this project would not be possible. RG is supported by NSERC, Amii, and CIFAR. RGK is supported by a Canada CIFAR AI Chair. MC is supported by a CIHR Health Science Impact Fellowship. This research was supported by a a DSI Catalyst Grant from the University of Toronto.
\end{acknowledgements}

\bibliography{main}
\title{Copula-Based Deep Survival Models for Dependent Censoring\\(Supplementary Material)}

\onecolumn
\raggedbottom
\maketitle
\tableofcontents
\vfill

\appendix

\section{Table of Notation}
\label{appx:table_of_notation}

\begin{table}[H]
    \setlength{\tabcolsep}{12pt}
    \begin{adjustbox}{center}
    \begin{tabular}{ll}
    $\textbf{1}^N$ & $N$-vector filled with 1's.\\
    $\mathbbm{1}[\cdot]$ & Indicator function.\\
    $\mathcal{L}(\cdot)$ & Likelihood function.\\
    $\ell(\cdot)$ & Log-likelihood function.\\
 $X \in \mathcal{X}$ & Covariates of one instance (as elements of the covariate space, $\mathcal{X}$).\\
 $T_E \in \mathbb{R}_+$ & Event time.\\
 $T_C \in \mathbb{R}_+$ & Censorship time.\\
 $T_{\text{obs}} \in \mathbb{R}_+$ & Time of last observation; the minimum of $T_E, T_C$.\\
 $T \in \mathbb{R}_+$ & Either event or censoring time; used in contexts where a quantity may refer to either.\\
 $\delta \in \{0,1\}$ & Event indicator. Equal to 1 if the observed time is the event time; 0 otherwise.\\
 $\mathcal{D} \subset \mathcal{X} \times \mathbb{R}_+ \times \{0,1\}$ & Survival dataset of the form $\{(X^{(i)}, T^{(i)}_\text{obs}, \delta^{(i)})\}_{i=1}^N$.\\
 $S_T \in \mathcal{S}$ & Survival function, $S : \mathbb{R}_+ \rightarrow [0,1]$, and space of survival functions, $\mathcal{S}$.\\
 $f_{T}$ & Probability density function over time, representing $\prob(T=t)$.\\
 $F_{T}$ & Cumulative density function over time, representing $\prob(T < t)$.\\
 $C$ & A copula. If written as $C_\theta$, this denotes a copula parameterized by the dependence parameter $\theta$.\\
 $u_1, u_2$ & Inputs to a copula function. It is assumed that these are uniformly distributed.   
\end{tabular}
\end{adjustbox}
\label{tab:notation}
\end{table}

\section{Copula Formulae and Algorithms}
\label{appx:copula_formulae_and_algorithms}

\subsection{Table of Preliminaries}
\label{appx:table_of_preliminaries}

\begin{table}[H]
    \small
    \begin{adjustbox}{center}
    \begin{tabular}{|c||c|c|c|}
         \hline
         Copula & $C_\theta(u_1, u_2)$ & $\Theta$ & \parbox{0.5cm} {\begin{align*} \frac{\partial}{\partial u_1} C_\theta(u_1, u_2) \end{align*}} \\
         \hline
         \hline
         Independence Copula & $u_1u_2$ & N/A & \parbox{3cm} {\begin{align*} u_2 \end{align*}}\\
         \hline
         Clayton Copula & $\left(\max \left(u_1^{-\theta} + u_2^{-\theta} - 1, 0\right)\right)^{-1/\theta}$ & $[-1, \infty)\backslash\{0\}$ & \parbox{3cm} {\begin{align*} 
         \begin{cases}
            \left(u_1^{-\theta} + u_2^{-\theta}-1\right)^{\frac{-\theta-1}{\theta}}u_2^{-\theta-1} \quad &u_1^{-\theta} + u_2^{-\theta} > 1\\
            0 &\text{otherwise}
         \end{cases}
         \end{align*}}\\
         \hline 
         Frank Copula & \parbox{3cm}{\begin{align*}
    \frac{-1}{\theta} \log \left(1+\frac{(\exp(-\theta u_1)-1)(\exp(-\theta u_2)-1)}{\exp(-\theta)-1}\right)
        \end{align*}} & $\mathbb{R}\backslash\{0\}$ & \parbox{3cm} {\begin{align*} 
        \frac{\exp(-\theta u_1)(\exp(-\theta u_2)-1)}{\exp(-\theta)-1}
         \end{align*}}\\
         \hline
    \end{tabular}
    \end{adjustbox}
    \caption{A table of formulas representing different classes of bivariate copulas used in our experiments. This table provides $C_\theta(u_1, u_2)$, the formula for the cumulative distribution function of the copula;
    $\Theta$, representing the family $\Theta$ from which valid $\theta$ may be drawn; and $\frac{\partial}{\partial u_1} C_\theta(u_1, u_2)$, representing the partial derivative of the copula with respect to its first parameter. Due to the symmetric nature of these copulas, one can readily find $\frac{\partial}{\partial u_2} C_\theta(u_1, u_2)$ from $\frac{\partial}{\partial u_1} C_\theta(u_1, u_2)$ by simply interchanging $u_1$, $u_2$ (hence, we only provide $\frac{\partial}{\partial u_1} C_\theta(u_1, u_2)$).}
    \label{tab:my_label}
\end{table}

\subsection{Sampling from a Copula}
\label{appx:sampling_from_a_copula}

Algorithm 2 requires that we draw samples from the Clayton and Frank copulas. To do so, we implement the copula sampling scheme from in the Python \href{https://www.statsmodels.org/stable/index.html}{\texttt{statsmodels}} package \citep{seabold2010statsmodels}.

\subsection{Quantile Density Visualizations}
\label{appx:quantile_density_visualizations}

\begin{figure}[H]
    \centering
    \includegraphics[width=\textwidth]{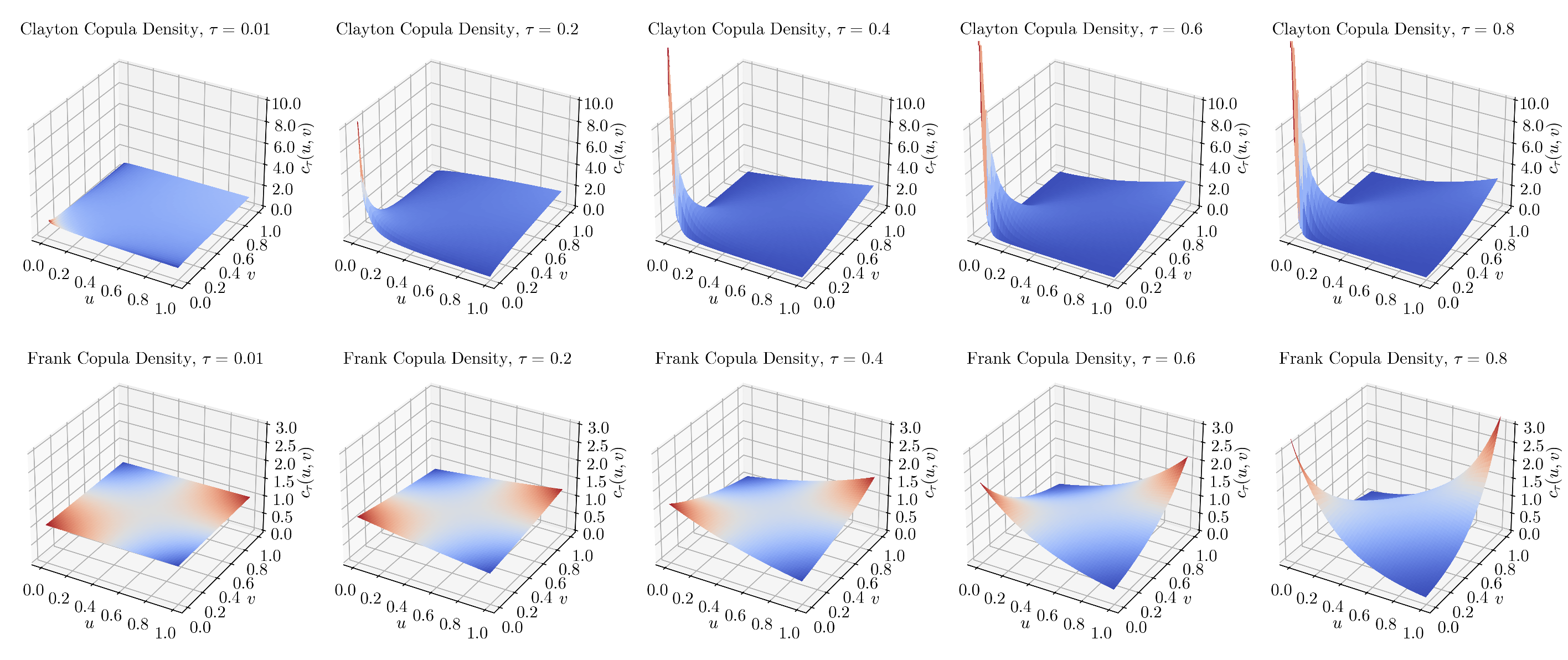}
    \caption{Plots of the densities for the Clayton (top row) and Frank (bottom row) copulas, under different degrees of dependence. These plots are functions of each of the copula's margins, $u$ and $v$. In practice, $u$ and $v$ are quantiles of the event and censoring distributions. Observe that, as the dependence increases, the difference in density between the on-diagonal points (points where $u \approx v$) and the off-diagonal points increases. Note also that, while the Clayton copula concentrates density around low quantiles (points where $u \approx v \approx 0$) as dependence increases, the Frank copula concentrates density more uniformly around the on-diagonal.}
    \label{fig:quantile-dependence}
\end{figure}

\subsection{Intuition for Copula Selection}
\label{appx:intuition_for_copula_selection}

In Section \ref{sec:discussion}, we discussed three different cases that can be used to build intuition around the forms of dependence induced by various copulas. In Figure \ref{fig:three_cases}, we visualize these cases, and relate them to the quantile density plots in Appendix \ref{appx:quantile_density_visualizations}. The point of this section is to build intuition regarding the \textit{a priori} selection of a copula, so we will necessarily make a few simplifications. For example, although the three cases we discuss are not exhaustive -- it is possible that the event and censoring survival curves cross (\textit{e.g.} if the event and censoring distributions have different baseline hazards) -- they present clean intuition relating the choice of copula to the structure of the joint density it produces.

\begin{figure}[H]
    \centering
    \includegraphics[width=0.9\textwidth]{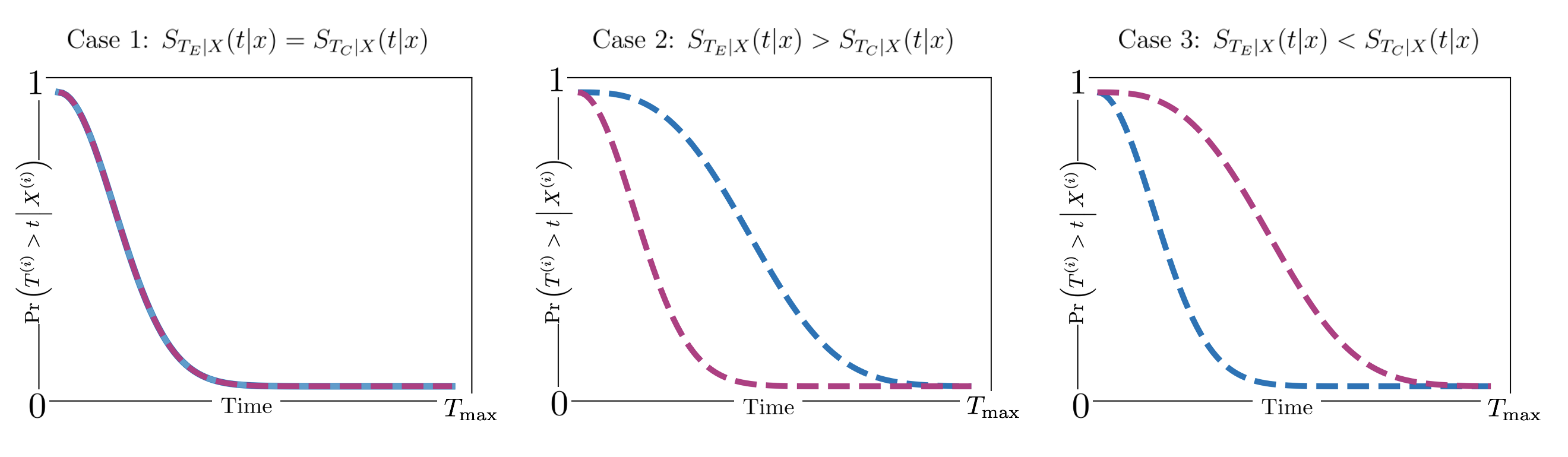}
    \caption{Three survival functions highlighting the three cases we presented in Section \ref{sec:discussion} of the main body. \textbf{Left}: the case where the conditional survival and censoring functions are the same. \textbf{Center}: the censoring survival function decays faster than the event survival function. \textbf{Right}: the event survival function decays faster than the censoring survival function. }
    \label{fig:three_cases}
\end{figure}

The key intuition for selecting a copula from domain knowledge can be drawn from Sklar's Theorem (Survival), which states that a joint distribution over event and censoring times can be modelled as two independent event and censoring distributions the quantiles of which are linked by a copula. When the event and censoring distributions are the same (left), the event quantile of a given time is the same as the censoring quantile for that same time. Thus, an increased dependence between event and censoring quantiles is directly reflected in a positive dependence between event and censoring times. When the censoring survival curve decays more quickly than the event survival curve, the event quantile of a given event time is higher than the censoring quantile for that same time. Therefore, increasing the dependence between event and censoring quantiles increases the likelihood that the censoring time precedes the event time. By symmetry, the opposite is true when the event survival curve decays more quickly than the censoring survival curve. An increase in dependence beteween quantiles in this setting increases the likelihood that the event itme precedes the censoring time under the model.

\section{Derivations}
\label{appx:derivations}

\subsection{The Right-Censored Likelihood}
\label{appx:the_right_censored_likelihood}

As a starting point for the subsequent derivations, we discuss the intuition behind the general likelihood for right-censored survival data (Equation~\ref{eq:generallikelihood}).

Recall that a survival dataset $\mathcal{D}$ consists of $N$ i.i.d. samples of the form 
$\{(X^{(i)}, T^{(i)}_\text{obs}, \delta^{(i)})\}_{i=1}^N \ \subset\ \mathcal{X} \times \mathbb{R}_+ \times \{0,1\}$.
The likelihood expressed in Equation \ref{eq:generallikelihood} uses the $\delta^{(i)}$ terms in the exponent as a conditional binary filter: raising a term to the power of $\delta^{(i)}$ ensures it is non-degenerate only when the patient experiences an event; raising a term to the power of $1-\delta^{(i)}$ ensures it is non-degenerate only when the patient is censored.

Let $f_{T_E, T_C | X}$ represent the joint density function of the event and censoring times, respectively, conditional on the patients' covariates. There are two mutually-exclusive, collectively-exhaustive into which we can decompose the right-censored likelihood for a given patient $i$:
\begin{enumerate}
\item\textbf{Case 1} ($\delta^{(i)} = 1$): If $\delta^{(i)} = 1$, the likelihood term should express that $T_E^{(i)} = T^{(i)}_\text{obs}$, and $T_C^{(i)} > T^{(i)}_\text{obs}$. This corresponds to the observation that the patient experienced the event at time $T^{(i)}_\text{obs}$, and was not censored prior to experiencing the event. The probability of this event under our density function is $\int_{T^{(i)}_\text{obs}}^\infty f_{T_E, T_C | X}(T^{(i)}_\text{obs}, t_c | X^{(i)})dt_c$.

\item\textbf{Case 2} ($\delta^{(i)} = 0$): If $\delta^{(i)} = 0$, the likelihood term should express that $T_C^{(i)} = T^{(i)}_\text{obs}$, and $T_E^{(i)} > T^{(i)}_\text{obs}$. This corresponds to the observation that the patient is censored at time $T^{(i)}_\text{obs}$, and did not experience an event prior to being censored. The probability of this event under our density function is $\int_{T^{(i)}_\text{obs}}^\infty f_{T_E, T_C | X}(t_e, T^{(i)}_\text{obs} | X^{(i)})dt_e$.
\end{enumerate}

Combining these two cases, and applying the assumption that our data is i.i.d., yields the general likelihood function for right-censored data.

\begin{equation}
\mathcal{L}(\mathcal{D}) = \prod_{i=1}^N \color{frenchblue}{\underbrace{\color{black}\left[\int_{T^{(i)}_\text{obs}}^\infty f_{T_E, T_C | X}(T^{(i)}_{\text{obs}},\, t_c\, |\, X^{(i)})\,dt_c\right]\color{frenchblue}}_{\Pr\left(T_E = T^{(i)}_{\text{obs}},\, T_C > T^{(i)}_{\text{obs}}\, |\, X^{(i)}\right)}}^{\color{black}\delta^{(i)}}\color{mediumred-violet}{\underbrace{\color{black}\left[\int_{T^{(i)}_{\text{obs}}}^\infty f_{T_E, T_C | X}(t_e,\, T^{(i)}_{\text{obs}}\, |\, X^{(i)})\,dt_e\right]}_{\color{mediumred-violet} \Pr\left(T_C = T^{(i)}_{\text{obs}},\, T_E > T^{(i)}_{\text{obs}}\, |\, X^{(i)}\,\right)}}^{\color{black}1-\delta^{(i)}}\color{black}
\label{eq:generallikelihood-supp}
\end{equation}

\subsection{The Right-Censored Log-Likelihood Under Conditional Independence}
\label{appx:the_right_censored_log_likelihood_under_conditional_independence}

Under the assumption that $T_E \perp T_C | X$, we can factorize the conditional density distributions in Equation \ref{eq:generallikelihood-supp}. $f_{T_E,T_C|X}$ factorizes into $f_{T_E|X}f_{T_C|X}$.
\begin{align}
\mathcal{L}(\mathcal{D}) &= \prod_{i=1}^N \left[f_{T_E|X}(T^{(i)}_\text{obs} | X^{(i)})\int_{T^{(i)}_\text{obs}}^\infty f_{T_C | X}(t_c | X^{(i)})dt_c\right]^{\delta^{(i)}} \left[f_{T_C|X}(T^{(i)}_\text{obs} | X^{(i)})\int_{T^{(i)}_\text{obs}}^\infty f_{T_E | X}(t_e | X^{(i)})dt_e\right]^{1-\delta^{(i)}}\\
&= \prod_{i=1}^N \left[f_{T_E|X}(T^{(i)}_\text{obs} | X^{(i)})\left(1-F_{T_C|X}(T^{(i)}_\text{obs}|X^{(i)})\right)\right]^{\delta^{(i)}} \left[f_{T_C|X}(T^{(i)}_\text{obs} | X^{(i)})\left(1-F_{T_E|X}(T^{(i)}_\text{obs}|X^{(i)})\right)\right]^{1-\delta^{(i)}}\\
&= \prod_{i=1}^N \left[f_{T_E|X}(T^{(i)}_\text{obs} | X^{(i)})S_{T_C|X}(T^{(i)}_\text{obs}|X^{(i)})\right]^{\delta^{(i)}} \left[f_{T_C|X}(T^{(i)}_\text{obs} | X^{(i)})S_{T_E|X}(T^{(i)}_\text{obs}|X^{(i)})\right]^{1-\delta^{(i)}}\\
\therefore \quad \ell(\mathcal{D}) &= \sum_{i=1}^N \delta^{(i)}\log\left[f_{T_E|X}(T^{(i)}_\text{obs} | X^{(i)})\right] + \delta^{(i)} \log \left[S_{T_C|X}(T^{(i)}_\text{obs}|X^{(i)})\right] + (1-\delta^{(i)}) \log
\left[f_{T_C|X}(T^{(i)}_\text{obs} | X^{(i)})\right] + \nonumber\\
& \qquad\quad(1-\delta^{(i)})\left[S_{T_E|X}(T^{(i)}_\text{obs}|X^{(i)})\right]
\label{eq:surv_indep}
\end{align}

\subsection{The Right-Censored Log-Likelihood Under Dependence Defined by a Copula}

\subsubsection{Proof of Lemma \ref{lemma:copula-conditional}}
\label{appx:proof_of_lemma_1}

\begin{manuallemma}{2}[Conditional Survival Function Under Sklar's Theorem (Survival)]
\label{lemma:copula-conditional-supp}
If $S_{T_E, T_C | X}(t_e, t_c | x) = \left.C(u_1, u_2)\middle|_{\substack{{u_1=S_{T_E|X}(t_e|x)}\\ {u_2=S_{T_C|X}(t_c|x)}}}\right.$, then, 
\begin{align}
    \int_{t_c}^\infty f_{T_C | T_E, X}(t_c | t_e, x) &= \frac{\partial}{\partial u_1} \left.C(u_1, u_2)\middle|_{\substack{{u_1=S_{T_E|X}(t_e|x)}\\ {u_2=S_{T_C|X}(t_c|x)}}}\right.
\end{align}
\end{manuallemma}
\begin{proof}
\begin{align}
    \int_{t_c}^\infty f_{T_C | T_E, X}(t_c | t_e, x) &= \frac{\int_{t_c}^\infty f_{T_C, T_E | X}(t_c, t_e | x) dt_c}{f_{T_E|X}(t_e|x)} && \text{(Def'n of Cond. Prob.)}\\
    &= \frac{\frac{-\partial}{\partial T_E} \int_{t_e}^\infty \int_{t_c}^\infty f_{T_C, T_E | X}(t_c, t_e | x) dt_c dt_e}{f_{T_E|X}(t_e|x)}\\
    &= \frac{\frac{-\partial}{\partial T_E} S_{T_C, T_E | X}(t_c, t_e | x)}{f_{T_E|X}(t_e|x)} &&\text{(Def'n of Survival Function)}\\
    &= \frac{\frac{-\partial}{\partial T_E} \left(C(u_1, u_2)\middle|_{\substack{{u_1=S_{T_E|X}(t_e|x)}\\ {u_2=S_{T_C|X}(t_c|x)}}}\right)}{f_{T_E|X}(t_e|x)} &&\text{(Sklar's Theorem)}\\
    &= \frac{\frac{-\partial}{\partial u_1} \left(C(u_1, u_2)\middle|_{\substack{{u_1=S_{T_E|X}(t_e|x)}\\ {u_2=S_{T_C|X}(t_c|x)}}}\right)\frac{\partial}{\partial T_E}S_{T_E|X}(t_e|x)}{f_{T_E|X}(t_e|x)} &&\text{(Chain Rule)}\\
    &= \frac{-\partial}{\partial u_1} \left(C(u_1, u_2)\middle|_{\substack{{u_1=S_{T_E|X}(t_e|x)}\\ {u_2=S_{T_C|X}(t_c|x)}}}\right)\cancelto{-1}{\frac{-f_{T_E|X}(t_e|x)}{f_{T_E|X}(t_e|x)}}\\
    &= \frac{\partial}{\partial u_1} \left(C(u_1, u_2)\middle|_{\substack{{u_1=S_{T_E|X}(t_e|x)}\\ {u_2=S_{T_C|X}(t_c|x)}}}\right)
\end{align}
\end{proof}

\textit{Corollary.} We can symmetrically apply this lemma to the converse case, $f_{T_E|T_C,X}$, to obtain:
\begin{equation}
    \int_{t_e}^\infty f_{T_E | T_C, X}(t_e | t_c, x) = \frac{\partial}{\partial u_2} \left(C(u_1, u_2)\middle|_{\substack{{u_1=S_{T_E|X}(t_e|x)}\\ {u_2=S_{T_C|X}(t_c|x)}}}\right)
\end{equation}

\subsubsection{Derivation of the Right-Censored Log Likelihood Under a Copula}
\label{appx:derivation_of_right_censored_log_likelihood_under_a_copula}

Having now proven Lemma \ref{lemma:copula-conditional}, we apply it to derive a likelihood function for survival prediction under dependent censoring. We use Equation \ref{eq:generallikelihood} as the starting point for our derivation.

\begin{align}
\mathcal{L}(\mathcal{D}) = \prod_{i=1}^N & \left[\int_{T^{(i)}_\text{obs}}^\infty f_{T_E, T_C | X}(T^{(i)}_\text{obs}, t_c | X^{(i)})\,dt_c\right]^{\delta^{(i)}} \left[\int_{T^{(i)}_\text{obs}}^\infty f_{T_E, T_C | X}(t_e, T^{(i)}_\text{obs} | X^{(i)})\,dt_e\right]^{1-\delta^{(i)}}\\
= \prod_{i=1}^N &\left[f_{T_E|X}(T^{(i)}_\text{obs}| X^{(i)})\int_{T^{(i)}_\text{obs}}^\infty f_{T_C | T_E, X}(t_c | T^{(i)}_\text{obs}, X^{(i)})\,dt_c\right]^{\delta^{(i)}}\times &&\text{(Chain Rule)}\\
&\left[f_{T_C|X}(T^{(i)}_\text{obs}| X^{(i)})\int_{T^{(i)}_\text{obs}}^\infty f_{T_E | T_C, X}(t_e | T^{(i)}_\text{obs}, X^{(i)})\,dt_e\right]^{1-\delta^{(i)}}\nonumber\\
= \prod_{i=1}^N &\left[f_{T_E|X}(T^{(i)}_\text{obs}| X^{(i)})\frac{\partial}{\partial u_1}\left(C(u_1, u_2)\middle\vert_{\substack{u_1 = S_{T_E|X}(T^{(i)}_\text{obs}|X^{(i)})\\u_2 = S_{T_C|X}(T^{(i)}_\text{obs})|X^{(i)}} }\right)\right]^{\delta^{(i)}}\times&&\text{(Lemma \ref{lemma:copula-conditional})}\\
&\left[f_{T_C|X}(T^{(i)}_\text{obs}| X^{(i)})\frac{\partial}{\partial u_2}\left(C(u_1, u_2)\middle\vert_{\substack{u_1 = S_{T_E|X}(T^{(i)}_\text{obs}|X^{(i)})\\u_2 = S_{T_C|X}(T^{(i)}_\text{obs})|X^{(i)}} }\right)\right]^{1-\delta^{(i)}}\nonumber\\
\therefore \quad \ell(\mathcal{D}) = \sum_{i=1}^N &\delta^{(i)}\log \left[f_{T_E|X}\left(T^{(i)}_\text{obs}| X^{(i)}\right)\right] + \delta^{(i)} \log \left[\frac{\partial}{\partial u_1}C(u_1, u_2)\middle\vert_{\substack{u_1 = S_{T_E|X}(T^{(i)}_\text{obs}|X^{(i)})\\u_2 = S_{T_C|X}(T^{(i)}_\text{obs})|X^{(i)}} }\right] +\\
&(1-\delta^{(i)})\log \left[f_{T_C|X}\left(T^{(i)}_\text{obs}| X^{(i)}\right)\right] + \nonumber\\
&(1-\delta^{(i)}) \log \left[\frac{\partial}{\partial u_2}C(u_1, u_2)\middle\vert_{\substack{u_1 = S_{T_E|X}(T^{(i)}_\text{obs}|X^{(i)})\\u_2 = S_{T_C|X}(T^{(i)}_\text{obs})|X^{(i)}} }\right]\nonumber
\end{align}

\subsection{The Weibull CoxPH Model}
\label{appx:the_weibull_coxph_model}

Recall that the Weibull CoxPH model is defined in terms of its hazard, as follows.
\begin{equation}
\label{eq:model-hazard}
h_{T|X}(\,t|X\,)\ =\ \left(\frac{\nu}{\rho}\right)\left(\frac{t}{\rho}\right)^{\nu-1} \exp\left(\,g_\psi(X)\,\right)
\end{equation}

Our method, however, relies on the ability to extract additional quantities -- the density ($\hat{f}_{T|X}$) and survival functions ($\hat{S}_{T|X}$) -- from the model, as these are essential to computing our likelihood function. In this section, we derive the closed-form expressions for these two quantities that are present in the main body of our work.

\subsubsection{The Survival Function}
\label{appx:the_survival_function}

The survival function under our model can be derived via its cumulative hazard.

\begin{definition}[Cumulative Hazard]
The \textit{cumulative hazard}
\begin{equation}
\hat{H}_{T|X}(t|X)\ \triangleq\ \int_{0}^t \hat{h}_{T|X}(u|X)du
\end{equation}
represents the integral of the hazard function over all time prior to a specified time, $t$.
\end{definition}

The cumulative hazard of the Weibull CoxPH can be expressed in closed form as follows:
\begin{align}
\label{eq:model-cum-hazard}
\hat{H}_{T|X}(\,t|X\,)\ &=\ \int_{0}^{t} \left(\frac{\nu}{\rho}\right)\left(\frac{u}{\rho}\right)^{\nu-1} \exp\left(\,g_\psi(X)\,\right)\,du \\
&= \ \left(\frac{t}{\rho}\right)^{\nu} \exp\left(\,g_\psi(X)\,\right)
\label{eq:cumhazard_closedform}
\end{align}
One alternative formulation of the survival function expresses $S_{T|X}$ in terms of the hazard function, as follows.
\begin{align}
\label{eq:model-survival}
S_{T|X}(t|X)\ \triangleq\ \exp(-H_{T|X}(t|X))
\end{align}
We can apply this identity to Equation \ref{eq:cumhazard_closedform} to obtain the following expression for $\hat{S}_{T|X}$ under the Weibull CoxPH model:
\begin{align}
\hat{S}_{T|X}(t|X)= \exp\left(-\left(\frac{t}{\rho}\right)^{\nu} \exp\left(\,g_\psi(X)\,\right)\right)
\end{align}

\subsubsection{The Density Function}
\label{appx:the_density_function}

From Equation 3, we know that the density of an event can be calculated as follows.
\begin{equation}
\label{eq:model-density}
f_{T|X}(t|X)\ =\ S_{T|X}(t|X) h_{T|X}(t|X)
\end{equation}

\subsection{A Stable Implementation}
\label{appx:a_stable_implementation}
In order to optimize a Weibull model in a stable way we used another representation of Weibull distribution. This new representation is derived by applying log transformation to the cumulative hazard function of Weibull distribution. 
\begin{equation}
    \begin{aligned}
         H_{T|X}(\,t|X\,) &= \exp(\log(H_{T|X}(t|X)))\\
    &=\exp\left(\log\left(\left(\frac{t}{\rho}\right)^{\nu} \exp(g_\psi(X))\right)\right)\\
    &=\exp(\nu\log(t) - \nu\log(\rho) + g_\psi(X))
    \end{aligned}
\end{equation}

Setting $\sigma = \frac{1}{\nu}$, $\mu = \log(\rho)$ ,and $f(x) = -\frac{g_\psi(X)}{\nu}$, gives us a long-cumulative hazard function of the following form.
    
\begin{equation}
\label{eq:model-cum-hazard_stable}
H_{T|X}(\,t|X\,)\ =\ \exp\left(\frac{\log(t) - \mu - f(x)}{\sigma}\right)
\end{equation}

\subsubsection{Hazard function}
\label{appx:hazard_function}

Given the formula for the cumulative hazard function we can derive the hazard function in the new format by taking the derivative of cumulative hazard with respect to $t$.
\begin{equation}
    h_{T|X}(\,t|X\,)=\ \frac{\partial H_{T|X}(\,t|X\,)}{\partial t} =\  \frac{H_{T|X}(\,t|X\,)}{t\sigma}
\end{equation}

\pagebreak
\section{Algorithms}
\label{appx:algorithms}

\subsection{Computing the Survival-$\ell_1$}
\label{appx:computing_the_survival_l1}

Here, we expand on the computation of the Survival-$\ell_1$ metric from the main paper by providing an algorithm for the explicit computation of the inner term of the Survival-$\ell_1$ metric, as well as the value $T_{\text{max}}$ for the given pair of survival curves, $S, \hat{S}$:

\begin{align}
\mathcal{C}_{\textit{Survival-}\ell_1}(S, \hat{S})\quad =\quad \sum_{i=1}^N &\frac{1}{N \times T_{\text{max}}^{(i)}} \underbrace{\int_{0}^{\infty} \left|S_{T\,|\,X}(t\,|\,X^{(i)}) - \hat{S}_{T\,|\,X}(t\,|\,X^{(i)})\right| dt}_{\text{Inner Term}}\nonumber
\end{align}

Although the integral in the $\mathcal{C}_{\textit{Survival-}\ell_1}$ is over an infinite domain, in this approximation, we consider only the simplified case wherein the upper bound of integration is $T_\text{max}$.

\RestyleAlgo{ruled}
\SetKwComment{Comment}{$\ $\# }{ }
\SetKwComment{Commentt}{\# }{ }
\setcounter{algocf}{2}
{
\begin{algorithm}[H]
\label{alg:optimization}
\KwIn{
\begin{enumerate}
    \item $S_1, S_2$: Survival curves to compare under the Survival-$\ell_1$ metric. Here, we assume $S_1$ is the ground-truth survival curve, and $S_2$ is the estimated curve.
    \item $Q_{\lVert\cdot\rVert}$: Normalizing quantile.
    \item $N_\text{steps}$: Number of discretization steps.
\end{enumerate}
}
\KwResult{
\begin{enumerate}
    \item $\Delta_{\text{total}}$: a discretized approximation of the integral $\int_{0}^{T_{\text{max}}} \left|S_{1}(t\,|\,X^{(i)}) - {S}_{2}(t\,|\,X^{(i)})\right| dt$.
    \item $T_\text{max}$: This is used as a normalization weight when computing the full expression for the Survival-$\ell_1$ metric.
\end{enumerate}
}
\hrulefill\\
$T_\text{max} \gets {S}^{-1}_{1}\left(Q_{\lVert\cdot\rVert}\right)$\;
$\Delta_\text{total} \gets 0$\\
\For{$i = 1,\, ...\,,\, N_\text{steps}$}{
    $\Delta_{i;S_1,S_2} \gets \frac{T_{\text{max}}}{N_{\text{steps}}} \times \ell_1\left[{S_1\left(\frac{i \times T_{\max}}{N_{\text{steps}}}\right)}, {S_2\left(\frac{i \times T_{\max}}{N_{\text{steps}}}\right)}\right]$\;
    $\Delta_{\text{total}} \gets \Delta_{\text{total}} + \Delta_{i;S_1,S_2}$\;
}
\Return{$\Delta_{\text{total}}, T_{\text{max}}$}
\caption{Discrete Approximation of the Inner Term of the Survival-$\ell_1$}
\end{algorithm}
}

\pagebreak
\subsection{Creating a Semi-Synthetic Dataset with Dependent Censoring}
\label{appx:creating_a_semi_synthetic_dataset_with_dependent_censoring}

We convert a regression dataset to a survival dataset with dependent censoring using the following algorithm.

\RestyleAlgo{ruled}
\SetKwComment{Comment}{$\ $\# }{ }
\SetKwComment{Commentt}{\# }{ }
\SetKwInOut{Dependencies}{Dependencies}
{
\begin{algorithm}[H]
\label{alg:datagenerating}
\KwIn{
\begin{enumerate}
    \item $\mathcal{D}_{\text{reg}} = \left\{X^{(i)}, Y^{(i)}\right\}_{i=1}^N \subseteq \mathcal{X} \times \mathbb{R}_+$. Regression dataset consisting of covariates and labels.
    \item $C_\theta: [0,1] \times [0,1] \rightarrow [0,1]$. A bivariate, uniparametric copula.
\end{enumerate}
}
\KwResult{
\begin{enumerate}
    \item $\mathcal{D}_{C, \theta} \subseteq \mathcal{X} \times \mathbb{R}_+ \{0,1\}$. Artificially censored version of $D_\text{reg}$ in which the joint distribution between $Y$ and $T_C$ is governed by the application of Sklar's Theorem to the copula $C_\theta$.
\end{enumerate}
}

\hrulefill\\

\Comment{Learn a Weibull CoxPH model based on the outcomes of the train set without any censoring}
$\hat{W}_E \gets \texttt{Weibull-Linear}(Y, X, \textbf{1}^N)$\;
${W}_C \gets {W}_E$\;

${{W}_C}.\nu \gets {{W}_C}.\nu / 0.6$ \Comment{Decreases the variance of the censoring distribution}
$T_{C} \gets \textbf{0}^N$\;
$\mathcal{D}_{C, \theta} = \emptyset$\;
\For{$i = 1,\, ...\,,\, N$}{
    $u_1^{(i)} \gets \hat{S}_{W_E}(Y^{(i)}); $\Comment{Obtain event quantile}
    $u_2^{(i)} \sim C_\theta(\cdot \,\mid\, u_1^{(i)}); $\Comment{Sample censoring quantile conditionally from the copula}
    $T_C^{(i)} \gets \hat{S}_{W_C}^{-1}(u_2^{(i)})$; \Comment{Obtain censoring time via inv. censoring survival function}
    $\mathcal{D}_{C, \theta} \gets \mathcal{D}_{C, \theta}\, \cup\, \{(X^{(i)}, \min\left(Y^{(i)}, T_C^{(i)}\right), \mathbbm{1}[Y^{(i)} \leq T_C^{(i)}])\}$\;
}
\Return{$\mathcal{D}_{C, \theta}$}\;
\caption{Semi-Synthetic Dataset Construction with Dependent Censoring}
\label{alg:semi-synthetic}
\end{algorithm}
}

\section{Additional Experimental Details}
\label{appx:additional_experimental_details}

\subsection{Evaluation Metrics Are Biased Under Dependence}
\label{appx:evaluation_metrics_are_biased_under_dependence}

For this experiment, we sampled 10,000 data points according to Algorithm \ref{alg:datagenerating} with $X \in \mathbb{R}^{N \times 10} \sim \mathcal{U}_{[0,1]}$, $\nu_E^* = 4, \rho_E^* = 17, \psi_E^*(X) = X_{1}^{2}+X_{2}^{2}$, $\nu_C^* = 3, \rho_C^* = 16, \psi_C^*(X) = \sum_{i=1}^{3}\beta_{C_{i}}X_{i}^{2}$, where $ \beta_C \in [0,1]^{10} \sim \mathcal{U}_{[0,1]}$.

\subsection{Implementation Details}
\label{appx:implementation_details}

We halted the learning algorithms if the validation loss failed to improve for a consecutive 3000 epochs. The \texttt{Linear-Risk} experiments were conducted without any form of regularization, whereas the \texttt{Nonlinear-Risk} experiments employed $\ell_2$ regularization with a coefficient of $\lambda=0.001$. For all experiments, the learning rate remained constant at $0.001$.

\section{Datasets and Processing}
\label{appx:datasets_and_processing}

\subsection{Steel Industry Energy Consumption (\texttt{STEEL}) Dataset}
\label{appx:steel_industry_energy_consumption_dataset}

The \texttt{STEEL} dataset \citep{ve2021efficient, sathishkumar2020energy, sathishkumar2020industry} is a regression dataset from the UCI Machine Learning Repository \citep{asuncion2007uci}, comprising 35,040 observations of of the power consumption of plants run by DAEWOO Steel Co. Ltd in Gwangyang, South Korea. The data includes 9 covariates (including day of the week, type of load (light/medium/heavy), CO$_2$ measurements in PPM, and leading/lagging reactive power measurements), and one outcome variable (the industry energy consumption, measured in kWh). For our semi-synthetic experiment, we used $70\%$ of the data as the train set, $15\%$ as the validation set, and $15\%$ as the test set.

\subsection{Airfoil Self-Noise (\texttt{AIRFOIL}) Dataset}
\label{appx:airfoil_self_noise_dataset}

The \texttt{Airfoil} dataset \citep{Dua:2019} is another regression dataset from the UCI Machine Learning Repository \citep{asuncion2007uci}. It comprises 1,503 observations obtained from aerodynamic and acoustic tests of two and three-dimensional airfoil blade sections conducted in an anechoic wind tunnel. The data includes 6 covariates (including frequency, angle of attack, chord length, free-stream velocity, suction side displacement thickness) and one outcome variable (scaled sound pressure level). For our semi-synthetic experiment, we used $70\%$ of the data as the train set, $15\%$ as the validation set, and $15\%$ as the test set.

\section{Additional Semi-Synthetic Experimental Results}
\label{appx:additional_semi_synthetic_experimental_results}

For the experiments in this section we used a Clayton copula to censor the dataset as described in Algorithm \ref{alg:semi-synthetic}.

\subsection{Semi-Synthetic Survival Regression on the \texttt{STEEL} Dataset}
\label{appx:additional_semi_synthetic_results_steel_dataset}

Below, we present the results of our survival regression on the test set of the \texttt{STEEL} dataset. 

\begin{table}[H]
    \small
    \begin{adjustbox}{center}
    \begin{tabular}{|c|c|c|c|c|}
    \hline
    & $\tau = 0.2$ & $\tau = 0.4$ & $\tau = 0.6$ & $\tau = 0.8$\\
    \hline
    Weibull CoxPH (No Censoring) & 0.513 & 0.513 & 0.513 & 0.513\\
    \hline
    Weibull CoxPH (Independence Assuming) & 0.333 & 0.309 & 0.324 & 0.341\\
    Weibull CoxPH (Dependent, \textbf{ours}) & 0.371 & 0.442 & 0.512 & 0.508\\
    \hline
    \end{tabular}
    \end{adjustbox}
    \caption{A table of $R^2$ values given by performing survival regression on the \texttt{STEEL} dataset under various degrees of dependence induced by Algorithm \ref{alg:semi-synthetic}. A higher $R^2$ indicates a better performing algorithm. The top row represents the performance of a Weibull CoxPH model trained on the regression data without censoring; this should indicate an upper bound on the performance of any survival model under censoring. We find that the performance of our approach, though below the theoretical upper bound, lies substantially above that of the independence-assuming approach.}
    \label{tab:steel-results}
\end{table}

\subsection{Semi-Synthetic Survival Regression on the \texttt{AIRFOIL} Dataset}
\label{appx:additional_semi_synthetic_results_airfoil_dataset}

Below, we present the results of our survival regression on the test set of the \texttt{AIRFOIL} dataset. 
\begin{table}[H]
    \small
    \begin{adjustbox}{center}
    \begin{tabular}{|c|c|c|c|c|}
    \hline
    &$\tau = 0.2$& $\tau = 0.4$& $\tau = 0.6$& $\tau = 0.8$\\
    \hline
      Weibull CoxPH (No Censoring)&$0.572$&$0.572$&$0.572$& 0.572\\
    \hline
    Weibull CoxPH (Independence Assuming)&$0.583$&$0.549$&$0.465$&$0.330$\\
    
    Weibull CoxPH (Dependent, \textbf{ours})&$0.580$&$0.564$&$0.507$&$0.484$\\
    \hline
\end{tabular}
    \end{adjustbox}
    \caption{A table of $R^2$ values given by performing survival regression on the \texttt{AIRFOIL} dataset under various degrees of dependence induced by Algorithm \ref{alg:semi-synthetic}. The top row represents the performance of a Weibull CoxPH model trained on the regression data without censoring; this should indicate an upper bound on the performance of any survival model under censoring. While performance of both methods degrades as dependence increases, we find that our method is better able to obtain higher values of $R^2$ than the independence-assuming model under greater degrees of dependence.}
    \label{tab:steel-results}
\end{table}

\end{document}